%% file: cvpr.tex
\documentclass[final]{cvpr}

\pdfoutput=1
\usepackage{times}
\usepackage{epsfig}
\usepackage{graphicx}
\usepackage{amssymb}
\usepackage{amsmath}
\usepackage{cite}
\usepackage{subfigure}
\usepackage{mathtools}
\usepackage{verbatim}
\usepackage{booktabs} 
\usepackage{multirow}
\usepackage{array}
\usepackage{footnote}
\usepackage{tabularx}
\usepackage{color}
\usepackage{colortbl}
\usepackage[linesnumbered,lined,vlined,ruled,commentsnumbered]{algorithm2e}
\usepackage{listings}
\usepackage{listing}
\usepackage{xcolor}
\usepackage{framed}

\usepackage[pagebackref=true,breaklinks=true,colorlinks,bookmarks=false]{hyperref}

\DeclareMathOperator*{\argmin}{arg\,min}

\definecolor{mygray}{gray}{0.6}
\definecolor{mygray-bg}{gray}{0.9}
\newcommand{\myparagraph}[1]{\vspace{0.1em}\noindent\textbf{#1}}

\newcommand{\redt}[1]{\textcolor[rgb]{1,0,0}{#1}}
\newcommand{\redtext}[1]{\textcolor[rgb]{1,0,0}{#1}}
\newcommand{\cotronlvsapce}{\vspace{-0.1cm}}
\newcommand{\cotronlcaptionvsapce}{\vspace{-0.1cm}}
\def\confYear{CVPR 2021}
\newcommand{\beginsupp}{%
        \setcounter{table}{0}
        \renewcommand{\thetable}{S\arabic{table}}%
        \setcounter{figure}{0}
        \renewcommand{\thefigure}{S\arabic{figure}}%
     }
\newcommand{\mycaptionsupp}[1]{{\textcolor{red}{#1}}}
     
\definecolor{codegreen}{rgb}{0,0.6,0}
\definecolor{codegray}{rgb}{0.5,0.5,0.5}
\definecolor{codepurple}{rgb}{0.58,0,0.82}
\definecolor{backcolour}{rgb}{0.95,0.95,0.92}

\lstdefinestyle{mystyle}{
    backgroundcolor=\color{backcolour},   
    commentstyle=\color{codegreen},
    keywordstyle=\color{magenta},
    numberstyle=\tiny\color{codegray},
    stringstyle=\color{codepurple},
    basicstyle=\ttfamily\footnotesize,
    breakatwhitespace=false,         
    breaklines=true,                 
    captionpos=b,                    
    keepspaces=true,                 
    numbers=left,                    
    numbersep=5pt,                  
    showspaces=false,                
    showstringspaces=false,
    showtabs=false,                  
    tabsize=2
}

\begin{document}

%%%%%%%%% TITLE
\title{Adaptive Aggregation Networks for Class-Incremental Learning}
\author{Yaoyao Liu$^{1}$ 
\quad Bernt Schiele$^{1}$ 
\quad Qianru Sun$^{2}$\\
\\
\small $^{1}$Max Planck Institute for Informatics, Saarland Informatics Campus\\
\small $^{2}$School of Computing and Information Systems, Singapore Management University\\
\small {\texttt{\{yaoyao.liu, schiele\}@mpi-inf.mpg.de}}  \quad  {\texttt{qianrusun@smu.edu.sg}}
}

\maketitle

\input{sections/0_abstract}
\input{sections/1_introduction}
\input{sections/2_related_work}
\input{sections/3_method}
\input{sections/4_experiment}
\input{sections/5_conclusions}

{\small
\bibliographystyle{ieee_fullname}
\bibliography{egbib}
}

\clearpage
\input{supplementary/sections/6_supplementary.tex}

\end{document}

%% file: sections/0_abstract.tex
\begin{abstract}
Class-Incremental Learning (CIL) aims to learn a classification model with the number of classes increasing phase-by-phase. An inherent problem in CIL is the stability-plasticity dilemma between the learning of old and new classes, i.e., high-plasticity models easily forget old classes, but high-stability models are weak to learn new classes. We alleviate this issue by proposing a novel network architecture called Adaptive Aggregation Networks (AANets) in which we explicitly build two types of residual blocks at each residual level (taking ResNet as the baseline architecture): a stable block and a plastic block. We aggregate the output feature maps from these two blocks and then feed the results to the next-level blocks. We adapt the aggregation weights in order to balance these two types of blocks, i.e., to balance stability and plasticity, dynamically. We conduct extensive experiments on three CIL benchmarks: CIFAR-100, ImageNet-Subset, and ImageNet, and show that many existing CIL methods can be straightforwardly incorporated into the architecture of AANets to boost their performances\footnote{Code: \href{https://class-il.mpi-inf.mpg.de/}{https://class-il.mpi-inf.mpg.de/}}.
\end{abstract}

%% file: sections/1_introduction.tex
\section{Introduction}
\label{sec1}

AI systems are expected to work in an incremental manner when the amount of knowledge increases over time. They should be capable of learning new concepts while maintaining the ability to recognize previous ones.
However, deep-neural-network-based systems often suffer from serious forgetting problems (called ``catastrophic forgetting'') when they are continuously updated using new coming data. This is due to two facts: (i) the updates can override the knowledge acquired from the previous data~\cite{mccloskey1989catastrophic, McRae1993Catastrophic, Ratcliff1990catastrophic, shin2017continual, KemkerMAHK18}, and (ii) the model can not replay the entire previous data to regain the old knowledge.

To encourage solving these problems, \cite{rebuffi2017icarl} defined a class-incremental learning (CIL) protocol for image classification where the training data of different classes gradually come phase-by-phase. 
In each phase, the classifier is re-trained on new class data, and then evaluated on the test data of both old and new classes.
To prevent trivial algorithms such as storing all old data for replaying, 
there is a strict memory budget due to which a tiny set of exemplars of old classes can be saved in the memory.
This memory constraint causes a serious data imbalance problem between old and new classes, and indirectly causes the main problem of CIL -- the stability-plasticity dilemma~\cite{mermillod2013stability}.
In particular, higher plasticity results in the forgetting of old classes~\cite{mccloskey1989catastrophic},
while higher stability weakens the model from learning the data of new classes (that contain a large number of samples).
Existing CIL works try to balance stability and plasticity using data strategies. For example, as illustrated in Figure~\ref{fig_new_teaser} (a) and (b), some early methods train their models on the imbalanced dataset where there is only a small set of exemplars for old classes~\cite{rebuffi2017icarl,Li18LWF}, and recent methods include a fine-tuning step using a balanced subset of exemplars sampled from all classes~\cite{Castro18EndToEnd,hou2019lucir,douillard2020podnet}.
However, these data strategies are still limited in terms of effectiveness.
For example, when using the models trained after $25$ phases, LUCIR~\cite{hou2019lucir} and Mnemonics~\cite{liu2020mnemonics} ``forget'' the initial $50$ classes by $30\%$ and $20\%$, respectively, on the ImageNet dataset~\cite{russakovsky2015imagenet}.

In this paper, we address the stability-plasticity dilemma by introducing a novel network architecture called Adaptive Aggregation Networks (AANets). Taking the ResNet~\cite{He_CVPR2016_ResNet} as an example of baseline architectures, we explicitly build two residual blocks (at each residual level) in AANets: one for maintaining the knowledge of old classes (i.e., the stability) and the other for learning new classes (i.e., the plasticity), as shown in Figure~\ref{fig_new_teaser}~(c).
We achieve these by allowing these two blocks to have different levels of learnability, i.e., less learnable parameters in the stable block but more in the plastic one. 
We apply aggregation weights to the output feature maps of these blocks, sum them up, and pass the result maps to the next residual level.
In this way, we are able to dynamically balance the usage of these blocks by updating their aggregation weights. To achieve auto-updating, we take the weights as hyperparameters and optimize them in an end-to-end manner~\cite{finn2017model, Wu2019LargeScale, liu2020mnemonics}.

\input{figures/1_framework}

Technically, 
the overall optimization of AANets is bilevel. Level-1 is to learn the network parameters for two types of residual blocks, and level-2 is to adapt their aggregation weights. 
More specifically, level-1 is the standard optimization of network parameters, for which we use all the data available in the phase.
Level-2 aims to balance the usage of the two types of blocks, for which we optimize the aggregation weights using a balanced subset (by downsampling the data of new classes), as illustrated in Figure~\ref{fig_new_teaser}~(c).
We formulate these two levels in a bilevel optimization program (BOP)~\cite{SinhaMD18bilevelreview} that solves two optimization problems alternatively, i.e., update network parameters with aggregation weights fixed, and then switch.
For evaluation, we conduct CIL experiments on three widely-used benchmarks, CIFAR-100, ImageNet-Subset, and ImageNet. We find that many existing CIL methods, e.g., iCaRL~\cite{rebuffi2017icarl}, LUCIR~\cite{hou2019lucir}, Mnemonics Training~\cite{liu2020mnemonics}, and PODNet~\cite{douillard2020podnet}, can be directly incorporated in the architecture of AANets, yielding consistent performance improvements. We observe that a straightforward plug-in causes memory overheads, e.g., $26\%$ and $15\%$ respectively for CIFAR-100 and ImageNet-Subset. For a fair comparison, we conduct additional experiments under the settings of zero overhead (e.g., by reducing the number of old exemplars for training AANets), and validate that our approach still achieves top performance across all datasets.

\textbf{Our contribution} is three-fold: 
1) a novel and generic network architecture called AANets specially designed for tackling the stability-plasticity dilemma in CIL tasks; 
2) a BOP-based formulation and an end-to-end training solution for optimizing AANets; 
and 3) extensive experiments on three CIL benchmarks by incorporating four baseline methods in the architecture of AANets.

%% file: figures/1_framework.tex
\begin{figure*}
%\vspace{-0.2cm}
\centering
\includegraphics[width=6.5in]{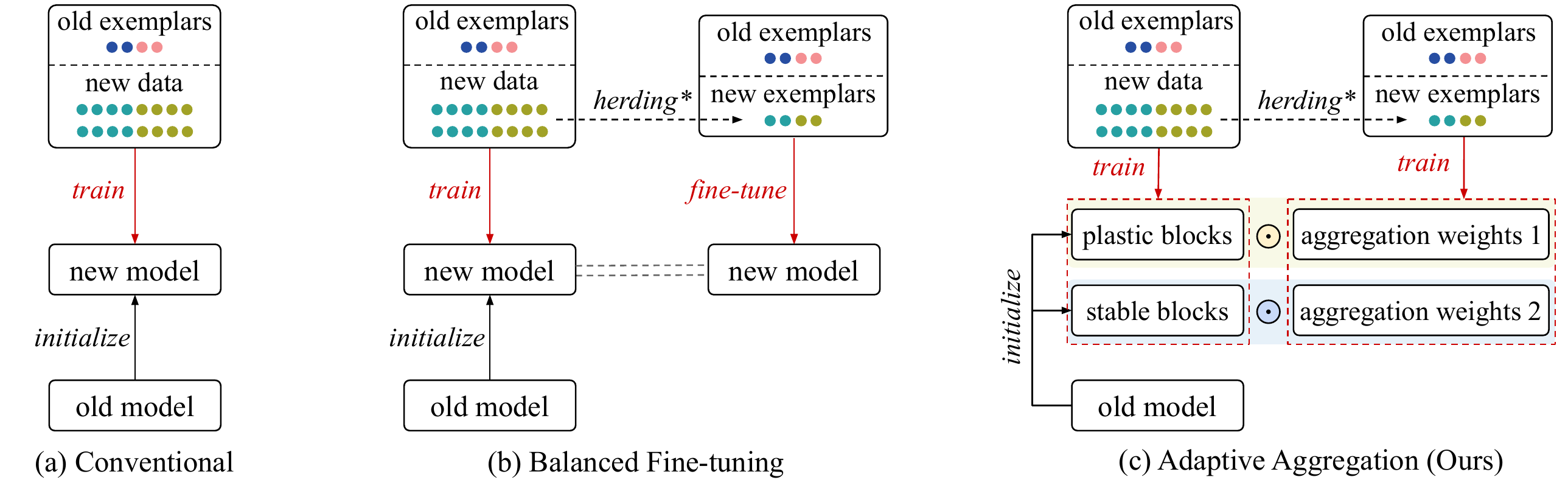}
%\cotronlcaptionvsapce
\vspace{0.1cm}
\caption{Conceptual illustrations of 
different CIL methods. 
(a) Conventional methods use all available data (which are imbalanced among classes) to train the model~\cite{rebuffi2017icarl,hou2019lucir}
(b) Recent methods~\cite{Castro18EndToEnd, hou2019lucir, douillard2020podnet, liu2020mnemonics} follow this convention but add a fine-tuning step on a balanced subset of all classes.
(c) The proposed Adaptive Aggregation Networks (AANets) is a new architecture and it applies a  different data strategy: using all available data to update the parameters of plastic and stable blocks, and the balanced set of exemplars to adapt the aggregation weights for these blocks. 
Our key lies in that adapted weights can balance the usage of the plastic and stable blocks, i.e., balance between plasticity and stability. 
*: \emph{herding} is the method to choose exemplars \cite{welling2009herding}, and can be replaced by others, e.g., \emph{mnemonics training} in \cite{liu2020mnemonics}. \textbf{We highlight that in the implementation of AANets, we strictly control the memory (i.e., the sizes of input data and residual blocks) within the same budget as the other methods. Please refer to the details in the section of experiments.}}
\label{fig_new_teaser}
\cotronlvsapce
\vspace{-0.3cm}
\end{figure*}

%% file: sections/2_related_work.tex
\section{Related Work}
\label{sec2}

\myparagraph{Incremental learning} aims to learn efficient machine models from the data that gradually come in a sequence of training phases. 
Closely related topics are referred to as continual learning~\cite{de2019continualsurvey,lopez2017gradient} and lifelong learning~\cite{chen2018lifelong,aljundi2017expert,li2019online}.
Recent incremental learning approaches are either task-based, i.e., all-class data come but are from a different dataset for each new phase~\cite{Li18LWF,shin2017continual,hu2018overcoming,chaudhry2018efficient,riemer2018learning,zhao2020maintaining,davidson2020sequential,chaudhry2018riemannian}, or class-based i.e., each phase has the data of a new set of classes coming from the identical dataset~\cite{rebuffi2017icarl,hou2019lucir,Wu2019LargeScale,Castro18EndToEnd,liu2020mnemonics,hu2021cil,Zhang_2021_CVPR}. 
The latter one is typically called class-incremental learning (CIL), and our work is based on this setting. 
Related methods mainly focus on how to solve the problems of forgetting old data. Based on their specific methods, they can be categorized into three classes: regularization-based, replay-based, and parameter-isolation-based~\cite{de2019literaturereview,prabhu12356gdumb}. 
\\
\myparagraph{\emph{Regularization-based}} methods introduce regularization terms in the loss function to consolidate previous knowledge when learning new data. 
Li et al.~\cite{Li18LWF} proposed the regularization term of knowledge distillation~\cite{HintonVD15}.
Hou et al.~\cite{hou2019lucir} introduced a series of new regularization terms such as for less-forgetting constraint and inter-class separation to mitigate the negative effects caused by the data imbalance between old and new classes.
Douillard et al.~\cite{douillard2020podnet} proposed an effective spatial-
based distillation loss applied throughout the model and also a representation comprising multiple proxy vectors for each object class.
Tao et al.~\cite{Tao2020topology} built the framework with a topology-preserving loss to maintain the topology in the feature space.
Yu et al.~\cite{yu2020CVPRsemantic} estimated the drift of previous classes during the training of new classes. 
\\
\myparagraph{\emph{Replay-based}} methods store a tiny subset of old data, and replay the model on them (together with new class data) to reduce the forgetting. 
Rebuffi et al.~\cite{rebuffi2017icarl} picked the nearest neighbors to the average sample per class to build this subset. 
Liu et al.~\cite{liu2020mnemonics} parameterized the samples in the subset, and then meta-optimized them automatically in an end-to-end manner taking the representation ability of the whole set as the meta-learning objective. 
Belouadah et al.~\cite{belouadah2019il2m} proposed to leverage a second memory to store  statistics of old classes in rather compact formats. 
\\
\myparagraph{\emph{Parameter-isolation-based}} methods are used in task-based incremental learning (but not CIL). Related methods dedicate different model parameters for different incremental phases, to prevent model forgetting (caused by parameter overwritten). 
If no constraints on the size of the neural network is given, one can grow new branches for new tasks while freezing old branches.
Rusu et al.~\cite{rusu2016progressive} proposed ``progressive networks'' to integrate the desiderata of different tasks directly into the networks.
Abati et al.~\cite{abati2020conditional} equipped each convolution layer with task-specific gating modules that select specific filters to learn each new task. 
Rajasegaran et al.\cite{NIPS2019_9429} progressively chose the optimal paths for the new task while encouraging to share parameters across tasks. 
Xu et al.~\cite{xu2018reinforced} searched for the best neural network architecture for each coming task by leveraging reinforcement learning strategies. 
\textbf{Our differences} with these methods include the following aspects. We focus on class-incremental learning, and more importantly, our approach does not continuously increase the network size. We validate in the experiments that under a strict memory budget, our approach can surpass many related methods and its plug-in versions on these related methods can bring consistent performance improvements.

\myparagraph{Bilevel Optimization Program} 
can be used to optimize hyperparameters of deep models.
Technically, the network parameters are updated at one level and the key hyperparameters are updated at another level~\cite{stackelberg1952theory,Wang2018Distillation,goodfellow2014generative,zhang2020deepemd,Liu2020E3BM,li2019learning}.
Recently, a few bilevel-optimization-based approaches have emerged for tackling incremental learning tasks.
Wu et al.~\cite{Wu2019LargeScale} learned a bias correction layer for incremental learning models using a bilevel optimization framework. 
Rajasegaran et al.~\cite{rajasegaran2020itaml} incrementally learned new tasks while learning a generic model to retain the knowledge from all tasks.
Riemer et al.~\cite{riemer2018learning2learn} learned network updates that are well-aligned with previous phases, such as to avoid learning towards any distracting directions.
In our work, we apply the bilevel optimization program to update the aggregation weights in our AANets.

%% file: sections/3_method.tex
\section{Adaptive Aggregation Networks (AANets)}
\label{sec_method}
\input{figures/2_f_connection}

Class-Incremental Learning (CIL) usually assumes $(N+1)$ learning phases in total, i.e., one initial phase and $N$ incremental phases during which the number of classes gradually increases~\cite{hou2019lucir,liu2020mnemonics,douillard2020podnet,hu2021cil}. 
In the initial phase, data $\mathcal{D}_0$ is available to train the first model $\Theta_0$.
There is a strict memory budget in CIL systems, so after the phase, only a small subset of $\mathcal{D}_0$ (exemplars denoted as $\mathcal E_0$) can be stored in the memory and used as replay samples in later phases.
Specifically in the $i$-th ($i\geq1$) phase, we load the exemplars of old classes $\mathcal E_{0:i-1}=\{\mathcal E_0, \dots, \mathcal E_{i-1}\}$ to train model $\Theta_i$ together with new class data $\mathcal{D}_i$.
Then, we evaluate the trained model on the test data containing both old and new classes.
We repeat such training and evaluation through all phases.

The key issue of CIL
is that the models trained at new phases easily ``forget'' old classes.
To tackle this, we introduce a novel architecture called AANets.
AANets is based on a ResNet-type architecture, and each of its residual levels is composed of two types of residual blocks: a plastic one to adapt to new class data and a stable one to maintain the knowledge learned from old classes. 
The details of this architecture are elaborated in Section~\ref{subsec_MANets_arch}. 
The steps for optimizing AANets are given in Section~\ref{subsec_optimization_MANets}.

\subsection{Architecture Details}
\label{subsec_MANets_arch}

In Figure~\ref{fig_framework_connection}, we provide an illustrative example of our AANets with three residual levels. 
The inputs $x^{[0]}$ are the images and the outputs $x^{[3]}$ are the features used to train classifiers.
Each of our residual ``levels'' consists of two parallel residual ``blocks'' (of the original ResNet~\cite{He_CVPR2016_ResNet}): the \textcolor[rgb]{0.93,0.47,0.32}{orange} one (called plastic block) will have its parameters fully adapted to new class data, while the \textcolor[rgb]{0.2,0.475,0.71}{blue} one (called stable block) has its parameters partially fixed in order to maintain the knowledge learned from old classes.
After feeding the inputs to Level 1, we obtain two sets of feature maps respectively from two blocks, and aggregate them after applying the aggregation weights $\alpha^{[1]}$. 
Then, we feed the resulted maps to Level 2 and repeat the aggregation. 
We apply the same steps for Level 3. 
Finally, we pool the resulted maps obtained from Level 3 to train classifiers.
Below we elaborate the details of this dual-branch design as well as the steps for feature extraction and aggregation.

\myparagraph{Stable and Plastic Blocks.}
We deploy a pair of stable and plastic blocks at each residual level, aiming to balance between the plasticity, i.e., for learning new classes, and stability, i.e., for not forgetting the knowledge of old classes. 
We achieve these two types of blocks by allowing different levels of learnability, i.e., less learnable parameters in the stable block but more in the plastic.
We detail the operations in the following.
In any CIL phase, Let $\eta$ and $\phi$ represent the learnable parameters of plastic and stable blocks, respectively.
$\eta$ contains all the convolutional weights,
while $\phi$ contains only the neuron-level scaling weights~\cite{sun2019meta}. Specifically, these scaling weights are applied on the model $\theta_{\mathrm{base}}$ obtained in the $0$-th phase\footnote{Related work~\cite{hou2019lucir,douillard2020podnet,liu2020mnemonics} learned $\Theta_0$ in the $0$-th phase using half of the total classes. We follow the same way to train $\Theta_0$ and freeze it as $\theta_{\mathrm{base}}$.}.
As a result, the number of learnable parameters $\phi$ is much less than that of $\eta$. 
For example, when using $3\times3$ neurons in $\theta_{\mathrm{base}}$, the number of learnable parameters $\phi$ is only $\frac{1}{3\times 3}$ of the number of full network parameters (while $\eta$ has the full network parameters). We further elaborate on these in the following paragraph. 

\myparagraph{Neuron-level Scaling Weights.} 
For stable blocks, we learn its neuron parameters in the $0$-th phase and freeze them in the other $N$ phases.
In these $N$ phases,  we apply a small set of scaling weights $\phi$ at the neuron-level, i.e., each weight for scaling one neuron in $\theta_\mathrm{base}$.
We aim to preserve the structural pattern within the neuron and slowly adapt the knowledge of the whole blocks to new class data. 
Specifically, we assume the $q$-th layer of $\theta_\mathrm{base}$ contains $R$ neurons, so we have $R$ neuron weights as $\{W_{q,r}\}_{r=1}^R$. For conciseness, we denote them as $W_{q}$.
For $W_{q}$, we learn $R$ scaling weights denoted as $\phi_{q}$
Let $X_{q-1}$ and $X_q$ be the input and output feature maps of the $q$-th layer, respectively. 
We apply $\phi_q$ to $W_q$ as follows,
\begin{equation}\label{eq_T_operation_layer}
     X_q =(W_q\odot\phi_{q}) X_{q-1},
\end{equation}
where $\odot$ donates the element-wise multiplication. Assuming there are $Q$ layers in total, the overall scaling weights can be denoted as $\phi=\{\phi_{q}\}_{q=1}^Q$.

\myparagraph{Feature Extraction and Aggregation.}
We elaborate on the process of feature extraction and aggregation across all residual levels in the AANets, as illustrated in Figure~\ref{fig_framework_connection}.
Let $\mathcal{F}^{[k]}_{\mu}{(\cdot)}$ denote
the transformation function of the residual block parameterized as $\mu$ at the Level $k$.
Given a batch of training images $x^{[0]}$, we 
feed them to AANets to compute the
feature maps at the $k$-th level (through the stable and plastic blocks respectively) as follows,
\begin{equation}\label{eq_kth_two_branch}
     x_{\phi}^{[k]} = \mathcal{F}^{[k]}_{\phi\odot \theta_{\mathrm{base}}}(x^{[k-1]});
     \quad x_{\eta}^{[k]} = \mathcal{F}^{[k]}_{\eta}(x^{[k-1]}).
\end{equation}
The transferability (of the knowledge learned from old classes) is different
at different levels of neural networks~\cite{yosinski2014transferable}.
Therefore, it makes more sense to apply
different aggregation weights for different levels of residual blocks.
Let $\alpha^{[k]}_{\phi}$ and $\alpha^{[k]}_{\eta}$ 
denote the 
aggregation weights of the stable and plastic blocks, respectively, at the $k$-th level. 
Then, the
weighted sum of $x_{\phi}^{[k]}$ and $x_{\eta}^{[k]}$ can be derived as follows,
\begin{equation}\label{eq_kth_alpha_fusion}
     x^{[k]} = \alpha^{[k]}_{\phi} \cdot x_{\phi}^{[k]} + \alpha^{[k]}_{\eta} \cdot x_{\eta}^{[k]}.
\end{equation}
In our illustrative example in Figure~\ref{fig_framework_connection}, there are three pairs of weights to learn at each phase. Hence, it becomes increasingly challenging to choose these weights manually if multiple phases are involved.
In this paper, we propose an learning strategy to automatically adapt these weights, i.e., optimizing the weights for different blocks in different phases, see details in Section~\ref{subsec_optimization_MANets}.

\subsection{Optimization Steps}
\label{subsec_optimization_MANets}

In each incremental phase, we optimize two groups of learnable parameters in AANets: (a) the neuron-level scaling weights $\phi$ for the stable blocks and the convolutional weights 
$\eta$ on the plastic blocks;
(b) the feature aggregation weights $\alpha$. 
The former is for network parameters and the latter is for hyperparameters. 
In this paper, we formulate the overall optimization process as a bilevel optimization program (BOP)~\cite{goodfellow2014generative,liu2020mnemonics}.

\myparagraph{The Formulation of BOP.}
In AANets, the network parameters $[\phi, \eta]$ are trained using the aggregation weights $\alpha$ as hyperparameters. 
In turn, $\alpha$ can be updated when temporarily fixing network parameters $[\phi, \eta]$. 
In this way, the optimality of $[\phi, \eta]$
imposes a constraint on $\alpha$ and vise versa.
Ideally, in the $i$-th phase, the CIL system aims to learn the optimal $\alpha_i$ and $[\phi_i, \eta_i]$ that minimize the classification loss on all training samples seen so far, i.e., $\mathcal{D}_i\cup\mathcal{D}_{0:i-1}$, so the ideal BOP can be formulated as,
\begin{subequations}
\label{eq_global_bilevel_program_original}
\begin{align}\label{eq_global_bilevel_upper_A}
    &\min_{\alpha_i} \mathcal{L}(\alpha_i, \phi^*_i, \eta^*_i; \mathcal{D}_{0:i-1}\cup \mathcal{D}_{i}) \\
\label{eq_global_bilevel_lower_A}
    &\ \text{s.t.} \ [\phi_i^*, \eta^*_i] = \argmin_{[\phi_i, \eta_i]} \mathcal{L}(\alpha_i, \phi_i, \eta_i; \mathcal{D}_{0:i-1}\cup \mathcal{D}_i),
\end{align}
\end{subequations}
where $\mathcal{L}(\cdot)$ denotes the loss function, e.g., cross-entropy loss. Please note that for the conciseness of the formulation, we use $\phi_i$ to represent $\phi_i\odot \theta_{\mathrm{base}}$ (same in the following equations).
We call Problem~\ref{eq_global_bilevel_upper_A} and Problem~\ref{eq_global_bilevel_lower_A} the \emph{upper-level} and
\emph{lower-level} problems, respectively.
%where ``\emph{meta}'' means the optimization is for hyperparameters and ``\emph{base}'' for network parameters. 
 
\myparagraph{Data Strategy.}
To solve Problem~\ref{eq_global_bilevel_program_original}, we need to use $\mathcal{D}_{0:i-1}$. However, in the setting of CIL~\cite{rebuffi2017icarl,hou2019lucir,douillard2020podnet}, we cannot access $\mathcal{D}_{0:i-1}$ but only a small set of exemplars $\mathcal{E}_{0:i-1}$, e.g., $20$ samples of each old class.
Directly replacing $\mathcal{D}_{0:i-1}\cup \mathcal{D}_{i}$ with $\mathcal{E}_{0:i-1}\cup \mathcal{D}_{i}$ in Problem~\ref{eq_global_bilevel_program_original} will lead to the forgetting problem for the old classes.
To alleviate this issue, we propose a new data strategy in which we use different training data splits to learn different groups of parameters:
1) in the \emph{upper-level} problem, $\alpha_i$ is used to balance the stable and the plastic blocks, so we use the balanced subset to update it, i.e., learning $\alpha_i$ on $\mathcal{E}_{0:i-1}\cup \mathcal{E}_{i}$ adaptively;
2) in the \emph{lower-level} problem, $[\phi_i, \eta_i]$ are the network parameters used for feature extraction, so we 
leverage all the available data
to train them, i.e., base-training $[\phi_i, \eta_i]$ on $\mathcal{E}_{0:i-1}\cup \mathcal{D}_{i}$. 
Based on these, we can reformulate the ideal BOP in Problem~\ref{eq_global_bilevel_program_original} as a solvable BOP as follows,
\begin{subequations}
\label{eq_global_bilevel_program}
\begin{align}\label{eq_global_bilevel_upper}
    &\min_{\alpha_i} \mathcal{L}(\alpha_i, \phi^*_i, \eta^*_i; \mathcal{E}_{0:i-1}\cup \mathcal{E}_{i}) \\
\label{eq_global_bilevel_lower}
    &\ \text{s.t.} \ [\phi_i^*, \eta^*_i] = \argmin_{[\phi_i, \eta_i]} \mathcal{L}(\alpha_i, \phi_i, \eta_i; \mathcal{E}_{0:i-1}\cup \mathcal{D}_i),
\end{align}
\end{subequations}
where Problem~\ref{eq_global_bilevel_upper} is the \emph{upper-level} problem and Problem~\ref{eq_global_bilevel_lower} is the \emph{lower-level} problem we are going to solve.

\myparagraph{Updating Parameters.}
We solve Problem~\ref{eq_global_bilevel_program} by alternatively updating two groups of parameters ($\alpha_i$ and $[\phi, \eta]$) across epochs, e.g., if $\alpha_i$ is updated in the $j$-th epoch, then $[\phi, \eta]$ will be updated in the $(j+1)$-th epoch, until both of them converge.
Taking the $i$-th phase as an example, we initialize ${\alpha}_i, \phi_i, \eta_i$ with ${\alpha}_{i-1}, \phi_{i-1}, \eta_{i-1}$, respectively. 
Please note that $\phi_0$ is initialized with ones, following~\cite{sun2019meta,sun2019mtlj}; $\eta_0$ is initialized with $\theta_{\mathrm{base}}$; and ${\alpha}_0$ is initialized with $0.5$.
Based on our \textbf{Data Strategy}, we use all available data in the current phase to solve the \emph{lower-level} problem, i.e., training [$\phi_i$, $\eta_i$] as follows,
\begin{equation}\label{eq_model_update_1}
    [\phi_i, \eta_i] \gets  [\phi_i, \eta_i] - \gamma_1 \nabla_{[\phi_i, \eta_i]}\mathcal{L}(\alpha_i, \phi_i, \eta_i; \mathcal{E}_{0:i-1}\cup \mathcal{D}_i).
\end{equation}
Then, we use a balanced exemplar set
to solve the \emph{upper-level} problem, i.e., training ${\alpha}_i$ as follows,
\begin{equation}\label{eq_model_update_2}
    {\alpha}_i \gets {\alpha}_i - \gamma_2 \nabla_{{\alpha}_i}\mathcal{L}({\alpha}_i, \phi_i, \eta_i; \mathcal{E}_{0:i-1}\cup \mathcal{E}_i),
\end{equation}
where $\gamma_1$ and $\gamma_2$ are the \emph{lower-level} and \emph{upper-level} learning rates, respectively. 

\input{pseudocode/algo}

\subsection{Algorithm}
\label{subsec_algo}

In Algorithm~\ref{alg:manet}, we summarize the overall training steps of the proposed AANets in the $i$-th incremental learning phase (where $i \in [1,...,N]$). Lines 1-4 show the preprocessing including loading new data and old exemplars (Line 1), initializing the two groups of learnable parameters (Lines 2-3), and selecting the exemplars for new classes (Line 4). Lines 5-12 optimize alternatively between the network parameters and the Adaptive Aggregation weights. In specific, Lines 6-8 and Lines 9-11 execute the training for solving the \emph{upper-level} and \emph{lower-level} problems, respectively. Lines 13-14 update the exemplars and save them to the memory.

%% file: figures/2_f_connection.tex
\begin{figure*}
\centering
%\vspace{-0.2cm}
\includegraphics[width=6.7in]{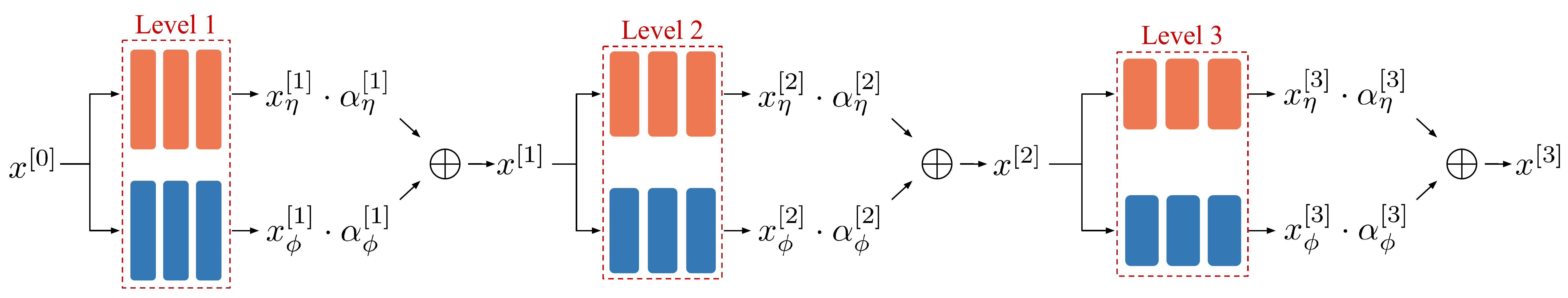}
\cotronlcaptionvsapce
\vspace{0.1cm}
\caption{
An example architecture of AANets with three levels of residual blocks.
At each level, we compute the feature maps from a stable block ($\phi\odot\theta_{\mathrm{base}}$, {\textcolor[rgb]{0.2,0.475,0.71}{blue}}) as well as a plastic block ($\eta$, {\textcolor[rgb]{0.93,0.47,0.32}{orange}}), respectively, aggregate the maps with adapted weights, and feed the result maps to the next level. The outputs of the final level are used to train classifiers. \textbf{We highlight that this is a logical architecture of AANets, and in real implementations, we strictly control the memory (i.e., the sizes of input data and residual blocks) within the same budget as related works which deploy plain ResNets. Please refer to the details in the section of experiments.}
}
\label{fig_framework_connection}
\cotronlvsapce
\vspace{-0.3cm}
\end{figure*}

%% file: pseudocode/algo.tex
  \begin{algorithm}
    \caption{AANets (in the $i$-th phase)} \label{alg:manet}
    \SetAlgoLined
    \SetKwInput{KwData}{Input}
    \SetKwInput{KwResult}{Output}
        \KwData{ New class data $\mathcal{D}_{i}$; old class exemplars $\mathcal{E}_{0:i-1}$; old parameters $\alpha_{i-1}$, $\phi_{i-1}$, $\eta_{i-1}$; base model $\theta_{\mathrm{base}}$.}
        \KwResult{new parameters $\alpha_{i}$, $\phi_{i}$, $\eta_{i}$; new class exemplars $\mathcal{E}_{i}$.}
    	
    	 Get $\mathcal{D}_{i}$ and load $\mathcal{E}_{0:i-1}$ from memory\;
    	 Initialize $[\phi_{i}, \eta_{i}]$ with $[\phi_{i-1}, \eta_{i-1}]$\;
    	 Initialize $\alpha_i$ with $\alpha_{i-1}$\;
    	 Select exemplars $\mathcal{E}_{i}\subsetneqq \mathcal{D}_{i}$, e.g. by herding~\cite{rebuffi2017icarl,hou2019lucir} or mnemonics training~\cite{liu2020mnemonics}\;
    		\For{\emph{epochs}}{
    		\For{\emph{mini-batches} \emph{\textbf{in}} $\mathcal{E}_{0:i-1}\cup \mathcal{D}_{i}$}{
    		 Train $[\phi_{i}, \eta_{i}]$ on $\mathcal{E}_{0:i-1}\cup \mathcal{D}_{i}$ by Eq.~\ref{eq_model_update_1}\;
            }  
            \For{\emph{mini-batches} \emph{\textbf{in}} $\mathcal{E}_{0:i-1}\cup \mathcal{E}_{i}$}{
            Learn ${\alpha}_i$ on $\mathcal{E}_{0:i-1}\cup \mathcal{E}_{i}$ by Eq.~\ref{eq_model_update_2}\;
            }
    		}
      Update exemplars $\mathcal{E}_i$, e.g. by herding~\cite{rebuffi2017icarl,hou2019lucir} or mnemonics training~\cite{liu2020mnemonics}\;
      Replace $\mathcal{E}_{0:i-1}$ with $\mathcal{E}_{0:i-1}\cup\mathcal{E}_i$ in the memory.
    \end{algorithm}

%% file: sections/4_experiment.tex
\section{Experiments}
\label{sec_exp}

We evaluate the proposed {AANets}
on three CIL benchmarks, i.e., CIFAR-100~\cite{krizhevsky2009learning}, ImageNet-Subset~\cite{rebuffi2017icarl} and ImageNet~\cite{russakovsky2015imagenet}. 
We incorporate AANets into four baseline methods and boost their model performances consistently for all settings. 
Below we describe the datasets and implementation details (Section~\ref{subsec_datasets}), followed by the results and analyses (Section~\ref{subsec_results}) which include a detailed ablation study, extensive comparisons to related methods, and some visualization of the results.

\subsection{Datasets and Implementation Details}
\label{subsec_datasets}
\input{tables/ablation_2}

\myparagraph{Datasets.}
We conduct CIL experiments on two datasets, CIFAR-100~\cite{krizhevsky2009learning} and ImageNet~\cite{russakovsky2015imagenet}, following closely related work~\cite{hou2019lucir,liu2020mnemonics,douillard2020podnet}. CIFAR-100 contains $60,000$ samples of $32\times32$ color images for $100$ classes. There are $500$ training and $100$ test samples for each class. ImageNet contains around $1.3$ million samples of $224\times224$ color images for $1000$ classes. There are approximately $1,300$ training and $50$ test samples for each class.
ImageNet is used in two CIL settings: one based on a subset of $100$ classes (ImageNet-Subset) and the other based on the full set of $1,000$ classes. 
The $100$-class data for {ImageNet-Subset} 
are sampled from ImageNet in the same way as~\cite{hou2019lucir,douillard2020podnet}.

\myparagraph{Architectures.}
Following the exact settings in~\cite{hou2019lucir,liu2020mnemonics}, we deploy a $32$-layer ResNet as the baseline architecture (based on which we build the AANets) for CIFAR-100.
This ResNet consists of $1$ initial convolution layer and $3$ residual blocks (in a single branch). Each block has $10$ convolution layers with $3\times3$ kernels. The number of filters starts from $16$ and is doubled every next block. After these $3$ blocks, there is an average-pooling layer to compress the output feature maps to a feature embedding. To build AANets, we convert these $3$ blocks into three levels of blocks and each level consists of a stable block and a plastic block, referring to Section~\ref{subsec_MANets_arch}.
Similarly, we build AANets for ImageNet benchmarks but taking an $18$-layer ResNet~\cite{He_CVPR2016_ResNet} as the baseline architecture~\cite{hou2019lucir,liu2020mnemonics}. 
Please note that there is no architecture change applied to the classifiers, i.e., using the same FC layers as in~\cite{hou2019lucir,liu2020mnemonics}.

\myparagraph{Hyperparameters and Configuration.}
The learning rates $\gamma_1$ and $\gamma_2$ are initialized as $0.1$ and $1\times10^{-8}$, respectively. 
We impose a constraint on each pair of $\alpha_{\eta}$ and $\alpha_{\phi}$ to make sure $\alpha_{\eta}+\alpha_{\phi}=1$.
For fair comparison, our training hyperparamters are almost the same as in~\cite{douillard2020podnet,liu2020mnemonics}.
Specifically, on the CIFAR-100 (ImageNet), we train the model for $160$ ($90$) epochs in each phase, and the learning rates are divided by $10$ after $80$ ($30$) and then after $120$ ($60$) epochs. 
We use an SGD optimizer with the momentum $0.9$ and the batch size $128$ to train the models in all settings.

\myparagraph{Memory Budget.} 
By default, we follow the same data replay settings used in~\cite{rebuffi2017icarl,hou2019lucir,liu2020mnemonics,douillard2020podnet}, where each time reserves $20$ exemplars per old class. 
In our ``strict memory budget'' settings, we strictly control the memory budget shared by the exemplars and the model parameters.
For example, if we incorporate AANets to LUCIR~\cite{hou2019lucir}, we need to reduce the number of exemplars to balance the additional memory used by model parameters (as AANets take around $20\%$ more parameters than plain ResNets).
As a result, we reduce the numbers of exemplars for AANets from $20$ to $13$, $16$ and $19$, respectively, for CIFAR-100, ImageNet-Subset, and ImageNet, in the ``strict memory budget'' setting.
For example, on CIFAR-100, we use $530k$ additional parameters, so we need to reduce $530k \mathrm{floats}\times4\mathrm{bytes/float}\div(32\times32\times3\mathrm{bytes}/\mathrm{image})\div100\mathrm{classes}\approx7\mathrm{images}/\mathrm{class}$.

\myparagraph{Benchmark Protocol.}
We follow the common protocol used in \cite{hou2019lucir,liu2020mnemonics,douillard2020podnet}.
Given a dataset, the model is trained on half of the classes in the $0$-th phase.
Then, 
it learns 
the remaining classes evenly in the subsequent $N$ phases. 
For $N$, there are three options as $5$, $10$, and $25$, and the corresponding settings are called ``$N$-phase''.
In each phase, the model is evaluated on the test data for all seen classes. The average accuracy (over all phases) is reported. 
For each setting, we run the experiment three times and report averages and $95\%$ confidence intervals.

\input{tables/sota}
\subsection{Results and Analyses}
\label{subsec_results}

Table~\ref{table:ablation} summarizes the statistics and results in $8$ ablative settings.
Table~\ref{table_sota} presents the results of $4$ state-of-the-art methods \emph{w/} and
\emph{w/o} AANets as a plug-in architecture, and the reported results from some other comparable work.
Figure~\ref{figure_gradcam} compares the activation maps (by Grad-CAM~\cite{selvaraju2017gradcam}) produced by different types of residual blocks and for the classes seen in different phases.
Figure~\ref{figure_values_plots} shows the changes of values of $\alpha_{\eta}$ and $\alpha_{\phi}$ across $10$ incremental phases.

\myparagraph{Ablation Settings.} 
Table~\ref{table:ablation} shows the ablation study.
By differentiating the numbers of learnable parameters, we can have $3$ block types: 1) ``all'' for learning all the convolutional weights and biases; 2) ``scaling'' for learning neuron-level scaling weights~\cite{sun2019meta} on the top of a frozen base model $\theta_{\mathrm{base}}$; and 3) ``frozen'' for using only $\theta_{\mathrm{base}}$ (always frozen). In Table~\ref{table:ablation}, the pattern of combining blocks is A+B where A and B stands for the plastic and the stable blocks, respectively.
Rows 1 is the baseline method LUCIR~\cite{hou2019lucir}.
Row 2 is a double-branch version for LUCIR without learning any aggregation weights.
Rows 3-5 are our AANets using different combinations of blocks.
Row 6-8 use ``all''+``scaling'' under an additional setting as follows. 1) Row 6 uses imbalanced data $\mathcal{E}_{0:i-1}\cup \mathcal{D}_{i}$ to train $\alpha$ adaptively. 2) Row 7 uses fixed weights $\alpha_{\eta}=\alpha_{\phi}=0.5$ at each residual level. 3) Row 8 is under the ``strict memory budget'' setting, where we reduce the numbers of exemplars to $14$ and $17$ for CIFAR-100 and ImageNet-Subset, respectively.

\myparagraph{Ablation Results.}
In Table~\ref{table:ablation}, if comparing the second block (ours) to the first block (single-branch and double-branch baselines), it is obvious that using AANets can clearly improve the model performance, e.g., ``scaling''+``frozen'' gains an average of $4.8\%$ over LUCIR for the ImageNet-Subset, by optimizing $1.4M$ parameters during CIL --- only $12.6\%$ of that in LUCIR.
Among Rows 3-5, we can see that for the ImageNet-Subset, models with the fewest learnable parameters (``scaling''+``frozen'') work the best.
We think this is because we use shallower networks for learning larger datasets (ResNet-32 for CIFAR-100; ResNet-18 for ImageNet-Subset), following the \textbf{Benchmark Protocol}.
In other words, $\theta_{\mathrm{base}}$ is quite well-trained with the rich data of half ImageNet-Subset ($50$ classes in the $0$-th phase), and can offer high-quality features for later phases. 
Comparing Row 6 to Row 3 shows the efficiency of using a balanced subset to optimize $\alpha$.
Comparing Row 7 to Row 3 shows the superiority of learning $\alpha$ (which is dynamic and optimal) over manually-choosing $\alpha$.
\\
\textbf{About the Memory Usage.}
By comparing Row 3 to Row 1, we can see that AANets can clearly improve the model performance while introducing small overheads for the memory, e.g., $26\%$ and $14.5\%$ on the CIFAR-100 and ImageNet-Subset, respectively.
If comparing Row~8 to Row~3, we find that though the numbers of exemplars are reduced (for Row~8), the model performance of AANets has a very small drop, e.g., only $0.3\%$ for the $5$-Phase CIL models of CIFAR-100 and ImageNet-Subset. %
Therefore, we can conclude that AANets can achieve rather satisfactory performance under strict memory control --- a desirable feature needed in class-incremental learning systems.

\input{figures/5_gradcam}
\input{figures/3_plot}

\myparagraph{Comparing to the State-of-the-Art.} 
Table~\ref{table_sota} shows that taking our AANets as a plug-in architecture for $4$ state-of-the-art methods~\cite{rebuffi2017icarl,hou2019lucir,liu2020mnemonics,douillard2020podnet} consistently
improves their model performances. E.g., for CIFAR-100,  LUCIR \emph{w/} AANets and Mnemonics \emph{w/} AANets respectively gains $4.9\%$ and $3.3\%$ improvements on average. 
From Table~\ref{table_sota}, we can see that our approach of using AANets achieves top performances in all settings.
Interestingly, we find that AANets can boost more performance for simpler baseline methods, e.g., iCaRL. iCaRL \emph{w/} AANets achieves mostly better results than those of LUCIR on three datasets, even though the latter method deploys various regularization techniques. 

\myparagraph{Visualizing Activation Maps.} Figure~\ref{figure_gradcam} demonstrates the activation maps visualized by Grad-CAM for the final model (obtained after $5$ phases) on ImageNet-Subset ($N$=5). 
The visualized samples from left to right are picked from the classes coming in the $0$-th, $3$-rd and $5$-th phases, respectively. 
For the $0$-th phase samples, the model makes the prediction according to foreground regions (right) detected by the stable block and background regions (wrong) by the plastic block.
This is because, through multiple phases of full updates, the plastic block forgets the knowledge of these old samples while the stable block successfully retains it. 
This situation is reversed when using that model to recognize the $5$-th phase samples. The reason is that the stable block is far less learnable than the plastic block, and may fail to adapt to new data.
For all shown samples, the model extracts features as informative as possible in two blocks. Then, it aggregates these features using the weights adapted from the balanced dataset, and thus can make a good balance of the features to achieve the best prediction.

\myparagraph{Aggregation Weights 
($\alpha_{\eta}$ and $\alpha_{\phi}$).
} Figure~\ref{figure_values_plots} shows the values of $\alpha_{\eta}$ and $\alpha_{\phi}$ learned during training $10$-phase models.
Each row displays three plots for three residual levels of AANets respectively. 
Comparing among columns, we can see that Level 1 tends to get larger values of $\alpha_{\phi}$, while Level 3 tends to get larger values of $\alpha_{\eta}$. This can be interpreted as lower-level residual blocks learn to stay stabler which is intuitively correct in deep models.
With respect to the learning activity of CIL models, it is to continuously transfer the learned knowledge to subsequent phases.
The features at different resolutions (levels in our case) have different transferabilities~\cite{yosinski2014transferable}.
Level 1 encodes low-level features that are more stable and shareable among all classes.
Level 3 nears the classifiers, and tends to be more plastic such as to fast to adapt to new classes.

%% file: tables/ablation_2.tex
\newcommand{\highestablation}[1]{\textbf{#1}}
\setlength{\tabcolsep}{1.10mm}{
\begin{table*}[htp]
  \small
  \centering
  \vspace{-0.2cm}
	  \begin{tabular}{cccccccccccccccccccc}
      \toprule
       \multirow{2.5}{*}{Row} && \multicolumn{3}{c}{\multirow{2.5}{*}{Ablation Setting}}  && \multicolumn{6}{c}{\emph{CIFAR-100} (acc.\%)} && \multicolumn{6}{c}{\emph{ImageNet-Subset} (acc.\%)}\\
       \cmidrule{7-12} \cmidrule{14-19} 
       && &   &  &&Memory&FLOPs& \#Param & $N$=5 & 10  & 25  && Memory&FLOPs&\#Param & $N$=5 & 10  & 25\\
       \midrule
        1 & \multicolumn{4}{c}{single-branch ``all''~\cite{hou2019lucir}} &&7.64MB&70M&469K& 63.17 & 60.14 & 57.54 &&330MB&1.82G&11.2M& 70.84 & 68.32 & 61.44 \\
        2  & \multicolumn{4}{c}{``all'' + ``all''} &&9.43MB&140M&938K&  64.49 & 61.89 & 58.87 &&372MB&3.64G&22.4M& 69.72 & 66.69 & 63.29\\
      \midrule
        3 & \multicolumn{4}{c}{``all'' + ``scaling''} &&9.66MB&140M&530K&  \highestablation{66.74} & \highestablation{65.29}
        & \highestablation{63.50} &&378MB&3.64G&12.6M& {{72.55}} &  {{69.22}} &  67.60 \\
        4 & \multicolumn{4}{c}{``all'' + ``frozen''} &&9.43MB&140M&469K&  65.62 & 64.05 & {63.67} &&372MB&3.64G&11.2M& 71.71 & 69.87 & 67.92\\
        5 & \multicolumn{4}{c}{``scaling'' + ``frozen''}  &&9.66MB&140M&60K&  64.71 & 63.65 & 62.89 &&378MB&3.64G&1.4M& \highestablation{73.01} & \highestablation{71.65} & \highestablation{70.30}\\
        \midrule
       6 & \multicolumn{4}{c}{\emph{w/o} balanced $\mathcal{E}$}  &&9.66MB&140M&530K&  65.91 & 64.70 & 63.08 &&378MB&3.64G&12.6M& 70.30 & 69.92 & 66.89\\
       7 & \multicolumn{4}{c}{\emph{w/o} adapted $\alpha$}  &&9.66MB&140M&530K& 65.89 & 64.49 & 62.89 &&378MB&3.64G&12.6M& 70.31 & 68.71 & 66.34\\
       8 & \multicolumn{4}{c}{strict memory budget}  &&7.64MB&140M&530K& 66.46 & 65.38 & 61.79 &&330MB&3.64G&12.6M& 72.21 & 69.10 & 67.10\\
      \bottomrule
    \end{tabular}
    \centering
    \cotronlcaptionvsapce
    \vspace{0.3cm}
	\caption{Ablation study. The baseline (Row 1) is LUCIR~\cite{hou2019lucir}. ``all'', ``scaling'', and ``frozen'' denote three types of blocks and they have different numbers of learnable parameters, e.g., ``all'' means all convolutional weights and biases are learnable. If we name them as A, B, and C, we use A+B in the table to denote the setting of using A-type and B-type blocks respectively as plastic and stable blocks. See more details in Section~\ref{subsec_results} \textbf{Ablation settings}. 
	Adapted ${\alpha}$ are applied on Rows~3-8. 
	``all''+``scaling'' is the default setting of Rows~6-8. 
	 ``\#Param'' indicates the number of learnable parameters. ``Memory'' denotes the peak memory for storing the exemplars and the learnable \& frozen network parameters during the model training through all phases. \emph{Please refer to more results in the supplementary materials.}
	}
	\label{table:ablation}
	\cotronlvsapce
	\vspace{-0.2cm}
    \end{table*}
    }

%% file: tables/sota.tex
\newcommand{\highest}[1]{\textbf{#1}}
\setlength{\tabcolsep}{1.0mm}{
\begin{table*}[ht]
  \small
  \centering
  \vspace{-0.2cm}
  \begin{tabular}{lccccccccccc}
  \toprule
   \multirow{2.5}{*}{Method} & \multicolumn{3}{c}{\emph{CIFAR-100}} && \multicolumn{3}{c}{\emph{ImageNet-Subset}} && \multicolumn{3}{c}{\emph{ImageNet}}\\
  \cmidrule{2-4} \cmidrule{6-8} \cmidrule{10-12}
   & $N$=5 & 10  & 25 && 5 & 10 & 25 && 5 & 10 & 25 \\
    \midrule
    \midrule
    LwF~\cite{Li18LWF} & 49.59 & 46.98 & 45.51 && 53.62 & 47.64 & 44.32 && 44.35 & 38.90 & 36.87\\
    BiC~\cite{Wu2019LargeScale} & 59.36 & 54.20 & 50.00 && 70.07 & 64.96 & 57.73 && 62.65 & 58.72 & 53.47 \\
    TPCIL~\cite{Tao2020topology} & 65.34 & 63.58 & -- && 76.27 & 74.81 & -- && 64.89 & 62.88 & -- \\
    \midrule
    \midrule
    iCaRL~\cite{rebuffi2017icarl} & 57.12\tiny{$\pm0.50$} & 52.66\tiny{$\pm0.89$} & 48.22\tiny{$\pm0.76$} && 65.44\tiny{$\pm0.35$} & 59.88\tiny{$\pm0.83$} & 52.97\tiny{$\pm1.02$} && 51.50\tiny{$\pm0.43$} & 46.89\tiny{$\pm0.35$} & 43.14\tiny{$\pm0.67$}  \\
    \cellcolor{mygray-bg}{\ \ \emph{w/} {AANets} (ours)} & \cellcolor{mygray-bg}{64.22}\tiny{$\pm0.42$} &  \cellcolor{mygray-bg}{60.26}\tiny{$\pm0.73$} &  \cellcolor{mygray-bg}{56.43}\tiny{$\pm0.81$} &\cellcolor{mygray-bg}{}&  \cellcolor{mygray-bg}{{73.45}}\tiny{$\pm0.51$} &  \cellcolor{mygray-bg}{71.78}\tiny{$\pm0.64$} &  \cellcolor{mygray-bg}{69.22}\tiny{$\pm0.83$} &\cellcolor{mygray-bg}{}&  \cellcolor{mygray-bg}{{63.91}}\tiny{$\pm0.59$} &  \cellcolor{mygray-bg}{{61.28}}\tiny{$\pm0.49$} &  \cellcolor{mygray-bg}{{56.97}}\tiny{$\pm0.86$} \\
    \midrule
    LUCIR~\cite{hou2019lucir}  & 63.17\tiny{$\pm0.87$} & 60.14\tiny{$\pm0.73$} & 57.54\tiny{$\pm0.43$} && 70.84\tiny{$\pm0.69$} & 68.32\tiny{$\pm0.81$} & 61.44\tiny{$\pm0.91$} && 64.45\tiny{$\pm0.32$} & 61.57\tiny{$\pm0.23$} & 56.56\tiny{$\pm0.36$} \\
    \cellcolor{mygray-bg}{\ \ \emph{w/} {AANets} (ours)}  & \cellcolor{mygray-bg}{66.74}\tiny{$\pm0.37$} & \cellcolor{mygray-bg}{65.29}\tiny{$\pm0.43$} & \cellcolor{mygray-bg}{\highest{63.50}}\tiny{$\pm0.61$} &\cellcolor{mygray-bg}{}&  \cellcolor{mygray-bg}{{72.55}}\tiny{$\pm0.67$} &  \cellcolor{mygray-bg}{{69.22}}\tiny{$\pm0.72$} &  \cellcolor{mygray-bg}{{67.60}}\tiny{$\pm0.39$} &\cellcolor{mygray-bg}{}&  \cellcolor{mygray-bg}{{64.94}}\tiny{$\pm0.25$} &  \cellcolor{mygray-bg}{{62.39}}\tiny{$\pm0.61$} &  \cellcolor{mygray-bg}{{60.68}}\tiny{$\pm0.58$} \\
    \midrule
    Mnemonics~\cite{liu2020mnemonics} & {{63.34}}\tiny{$\pm0.62$} & {{62.28}}\tiny{$\pm0.43$} &  {{60.96}}\tiny{$\pm0.72$} &&  {{72.58}}\tiny{$\pm0.85$} &  {{71.37}}\tiny{$\pm0.56$} &  {{69.74}\tiny{$\pm0.39$}} &&  {{64.54}}\tiny{$\pm0.49$} &  {{63.01}}\tiny{$\pm0.57$} &  {{61.00}}\tiny{$\pm0.71$} \\
    \cellcolor{mygray-bg}{\ \ \emph{w/} {AANets} (ours)} & \cellcolor{mygray-bg}{\highest{{67.59}}}\tiny{$\pm0.34$} & \cellcolor{mygray-bg}{\highest{{65.66}}}\tiny{$\pm0.61$} &  \cellcolor{mygray-bg}{{63.35}}\tiny{$\pm0.72$} &\cellcolor{mygray-bg}{}&  \cellcolor{mygray-bg}{72.91}\tiny{$\pm0.53$} &  \cellcolor{mygray-bg}{71.93}\tiny{$\pm0.37$} &  \cellcolor{mygray-bg}{70.70}\tiny{$\pm0.45$} &\cellcolor{mygray-bg}{}&  \cellcolor{mygray-bg}{{{65.23}}}\tiny{$\pm0.62$} &  \cellcolor{mygray-bg}{{{63.60}}}\tiny{$\pm0.71$} &  \cellcolor{mygray-bg}{{61.53}}\tiny{$\pm0.29$} \\
    \midrule
    PODNet-CNN~\cite{douillard2020podnet} & 64.83\tiny{$\pm1.11$} & 63.19\tiny{$\pm1.31$} &  60.72\tiny{$\pm1.54$} &&  75.54\tiny{$\pm0.29$} &  74.33\tiny{$\pm1.05$} &  68.31\tiny{$\pm2.77$} &&  66.95 &  64.13 & {{59.17}} \\
   \cellcolor{mygray-bg}{ \ \ \emph{w/} {AANets} (ours)}  & \cellcolor{mygray-bg}{66.31}\tiny{$\pm0.87$} & \cellcolor{mygray-bg}{64.31}\tiny{$\pm0.90$} &  \cellcolor{mygray-bg}{62.31}\tiny{$\pm1.02$} &\cellcolor{mygray-bg}{}& \cellcolor{mygray-bg}{ \highest{76.96}}\tiny{$\pm0.53$} &  \cellcolor{mygray-bg}{\highest{75.58}}\tiny{$\pm0.74$} &  \cellcolor{mygray-bg}{\highest{71.78}}\tiny{$\pm0.81$} &\cellcolor{mygray-bg}{}& \cellcolor{mygray-bg}{ {\highest{67.73}}\tiny{$\pm0.71$} } &  \cellcolor{mygray-bg}{{\highest{64.85}}}\tiny{$\pm0.53$} &  \cellcolor{mygray-bg}{\highest{{61.78}}\tiny{$\pm0.61$} } \\
  \bottomrule

\end{tabular}
\cotronlcaptionvsapce
\vspace{0.3cm}
  \caption{
  Average incremental accuracies (\%) of four state-of-the-art methods \emph{w/} and \emph{w/o} our AANets as a plug-in architecture. 
  In the upper block, we present some comparable results reported in some other related works.
{Please note 1) \cite{douillard2020podnet} didn't report the results for $N$=25 on the ImageNet, and we produce the results using their public code; 2) \cite{liu2020mnemonics} updated their results on arXiv (after fixing a bug in their code), different from its conference version; 3) for ``\emph{w/} AANets'', we use ``all''+``scaling'' blocks corresponding to Row~3 of Table~\ref{table:ablation}; 
and 4) 
\textbf{if applying ``strict memory budget'', there is little performance drop. Corresponding results are given in Table~\ref{table:ablation} and Table~\ref{table_sota_stric_memory} in the supplementary materials.}}
}
  \label{table_sota}
  \cotronlvsapce
  \vspace{-0.3cm}
\end{table*}
}

%% file: figures/5_gradcam.tex
\begin{figure*}
%\vspace{-0.2cm}
\newcommand{\newincludegraphics}[1]{\includegraphics[height=2.0in]{#1}}
\centering
\newincludegraphics{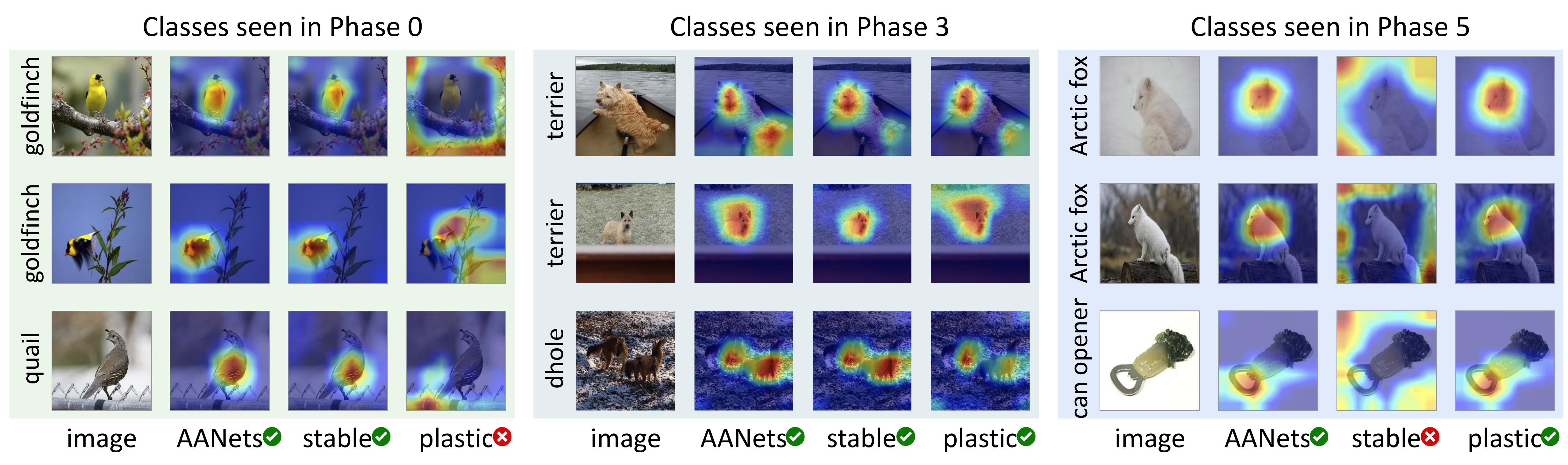}
\cotronlcaptionvsapce
\caption{The activation maps using Grad-CAM~\cite{selvaraju2017gradcam} for the 5-th phase (the last phase) model on ImageNet-Subset ($N$=5). 
Samples are selected from the classes coming in the 0-th phase (left), the 3-rd phase (middle), and the 5-th phase (right), respectively. 
Green tick (red cross) means the discriminative features are activated on the object regions successfully (unsuccessfully).
$\bar{\alpha}_{\eta}=0.428$ and $\bar{\alpha}_{\phi}=0.572$.
}
\label{figure_gradcam}
\vspace{-0.2cm}
\cotronlvsapce
\end{figure*}

%% file: figures/3_plot.tex
\begin{figure}
\centering

\newcommand{\plotincludegraphics}[1]{\includegraphics[height=1.0in]{#1}}

\newcommand{\newincludegraphicscifar}[1]{\includegraphics[height=1.1in]{#1}}
\newcommand{\newincludegraphics}[1]{\includegraphics[height=1.1in]{#1}}
\centering
\subfigure[CIFAR-100 ($N$=10)]{
\newincludegraphicscifar{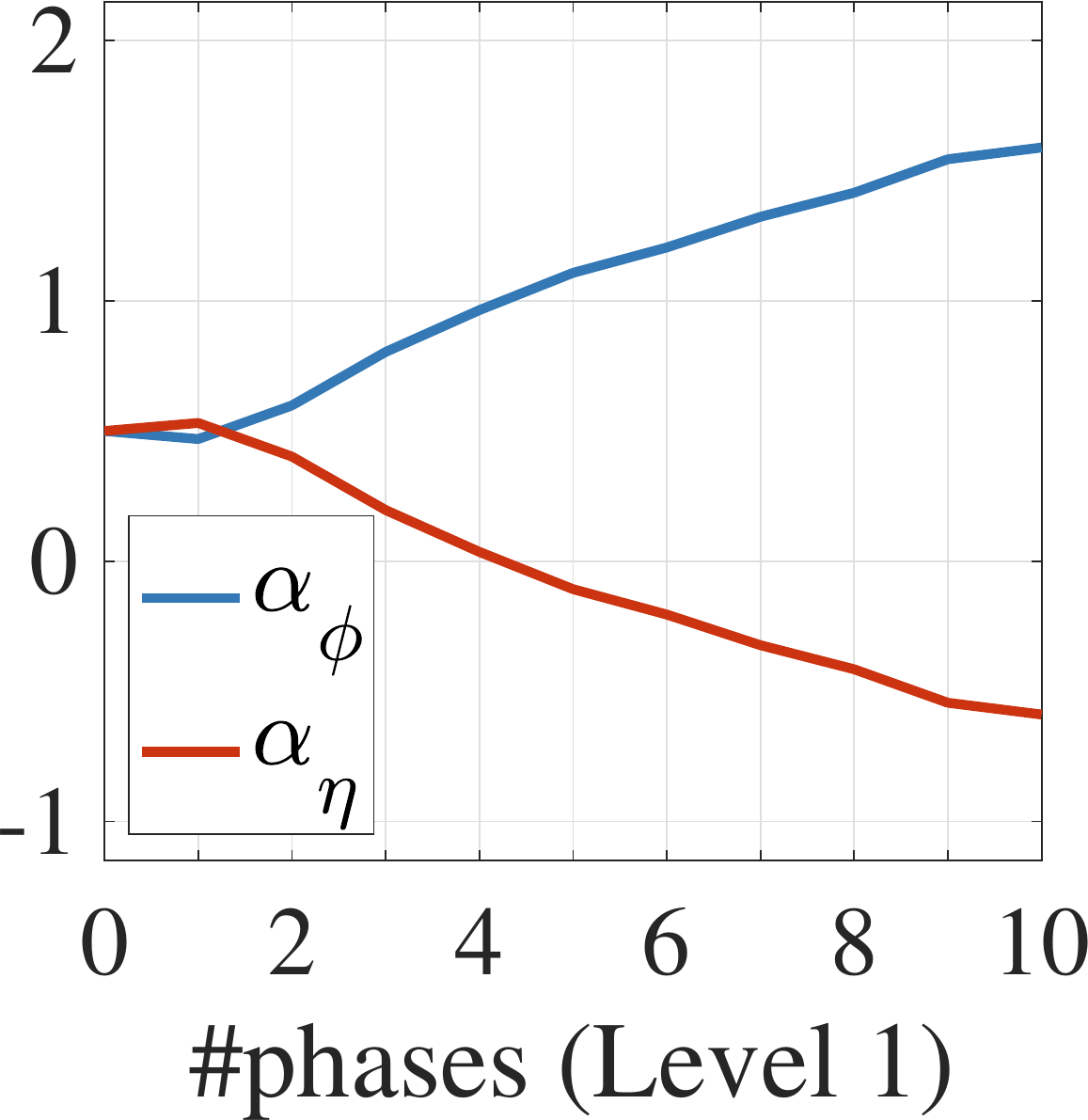}
\newincludegraphicscifar{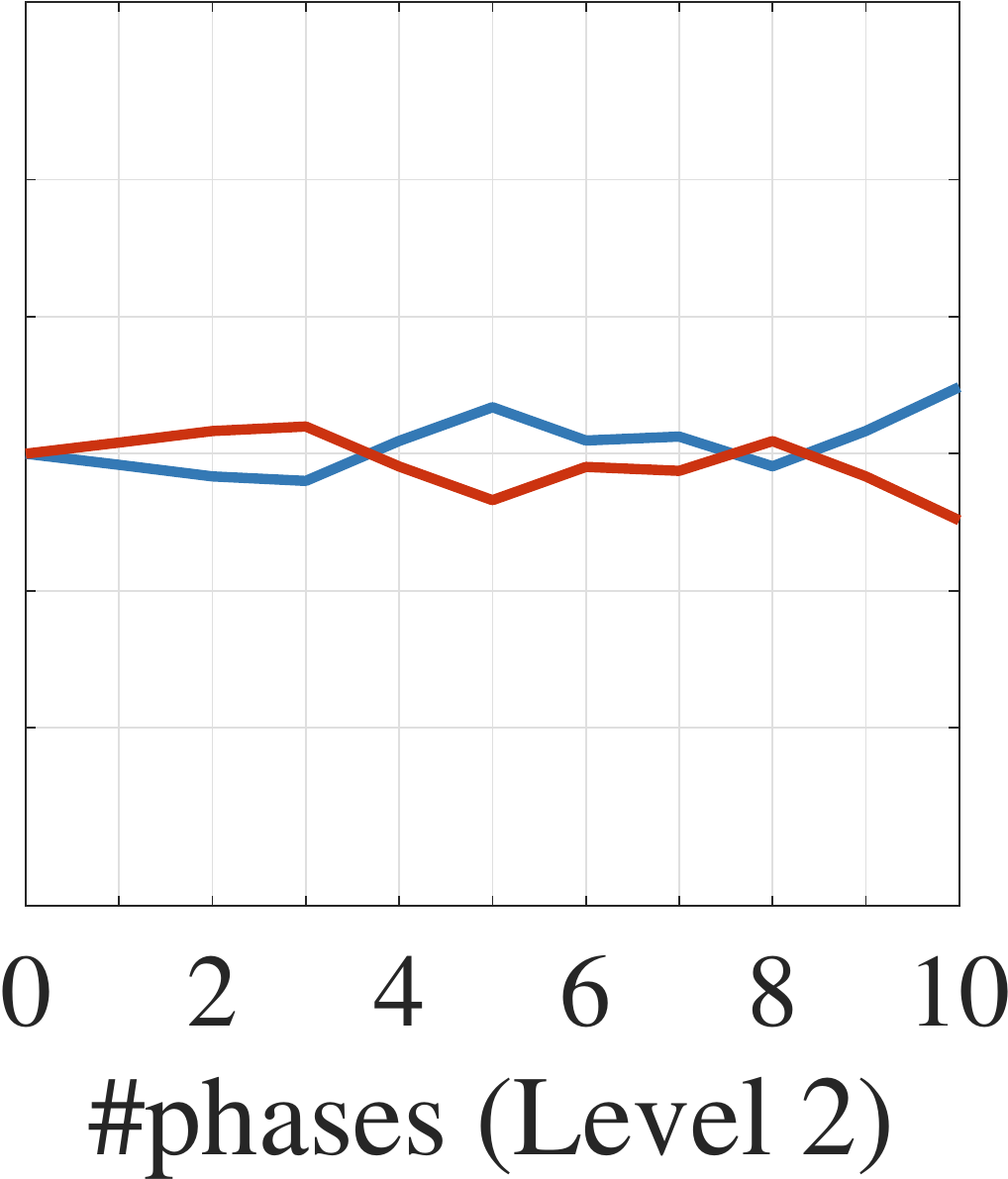}
\newincludegraphicscifar{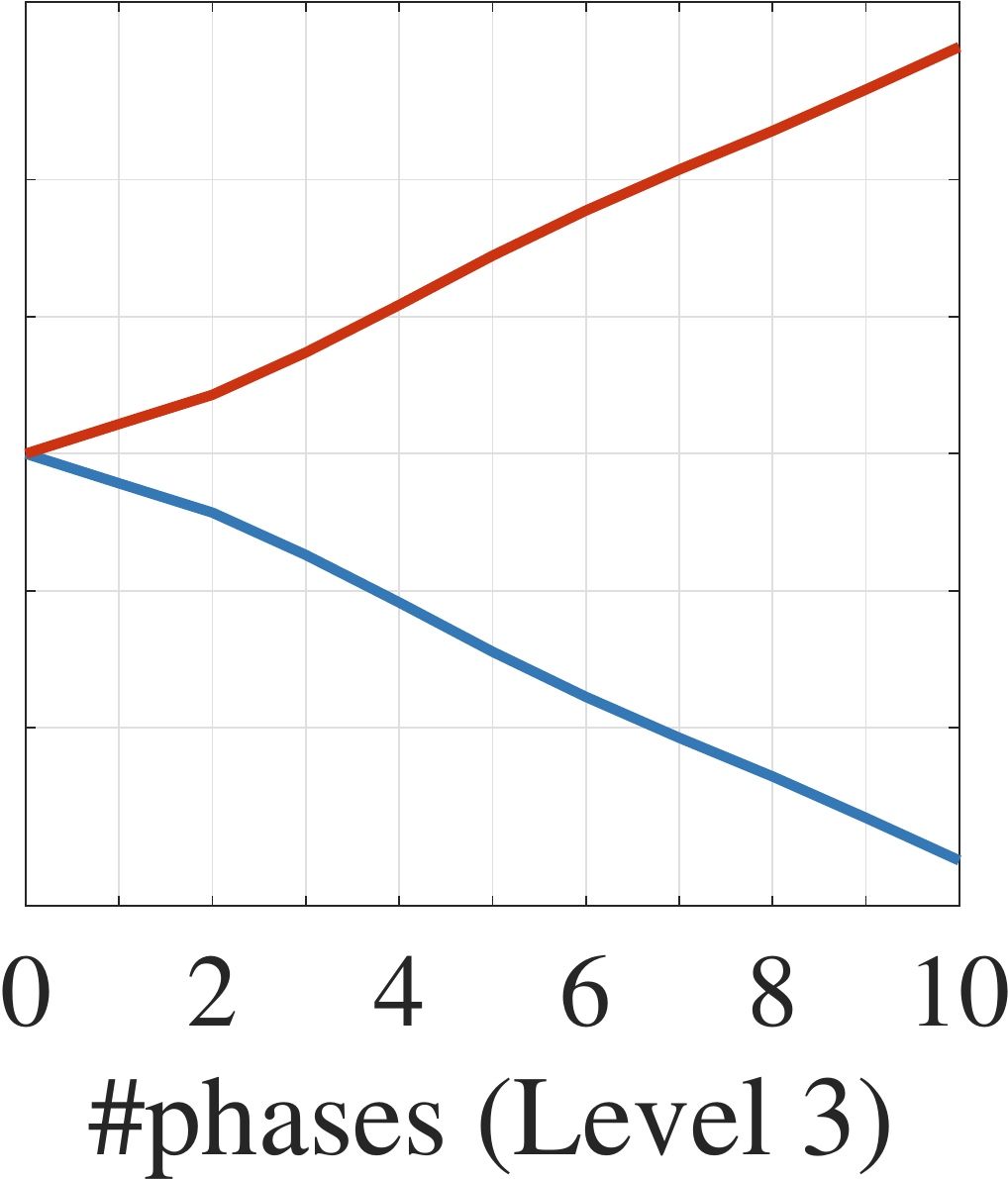}
}
\subfigure[ImageNet-Subset ($N$=10)]{
\newincludegraphics{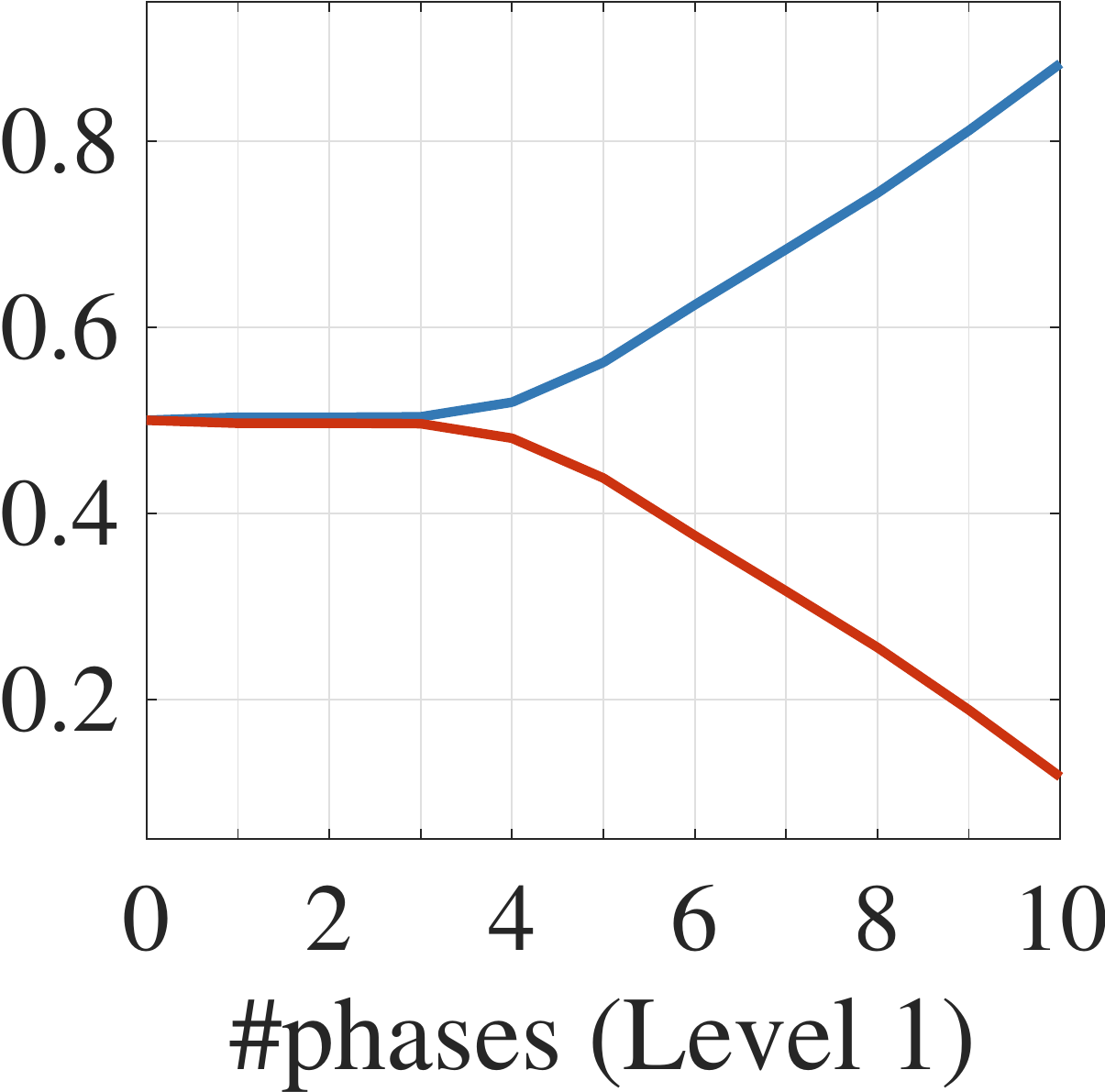}
\newincludegraphics{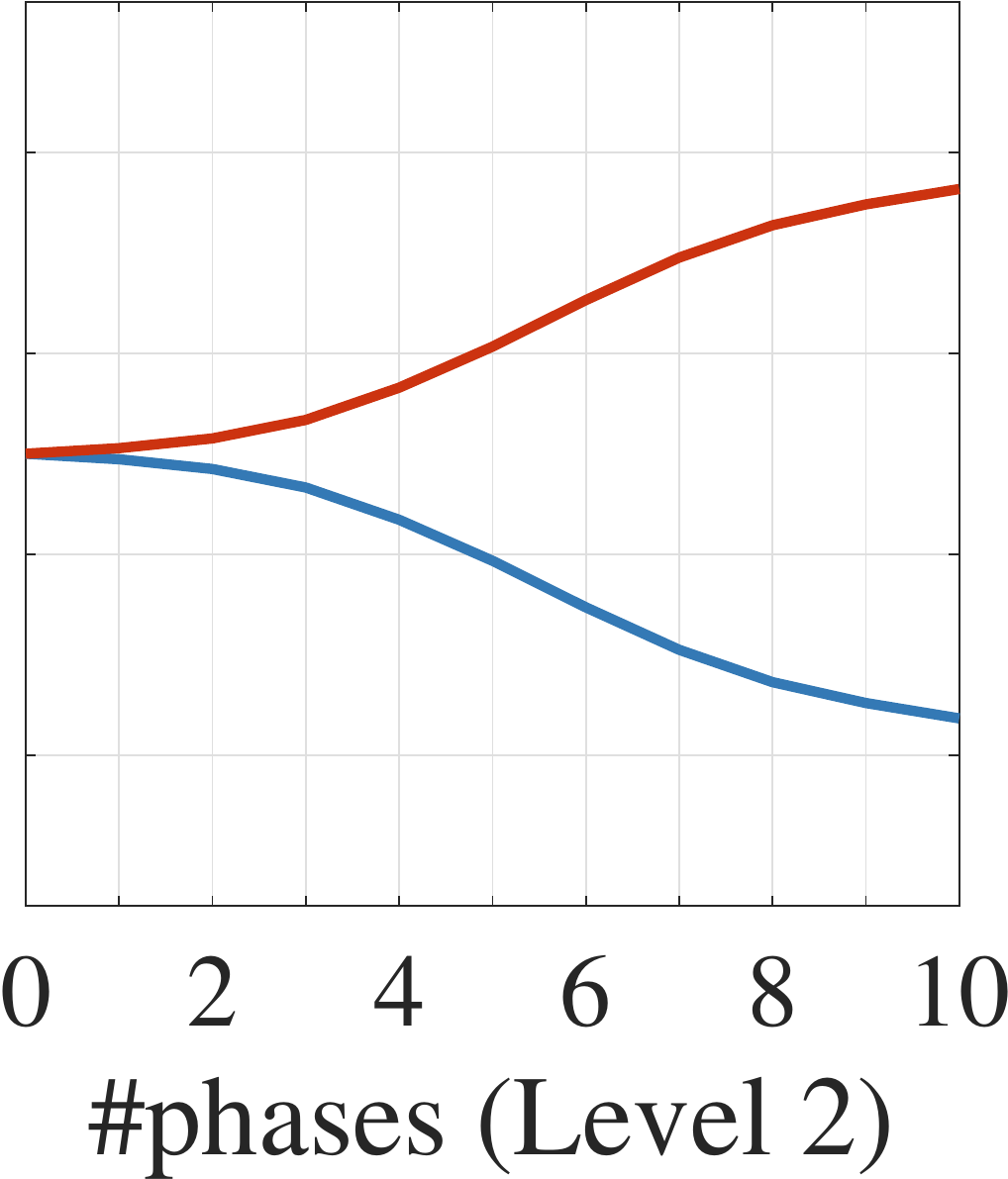}
\newincludegraphics{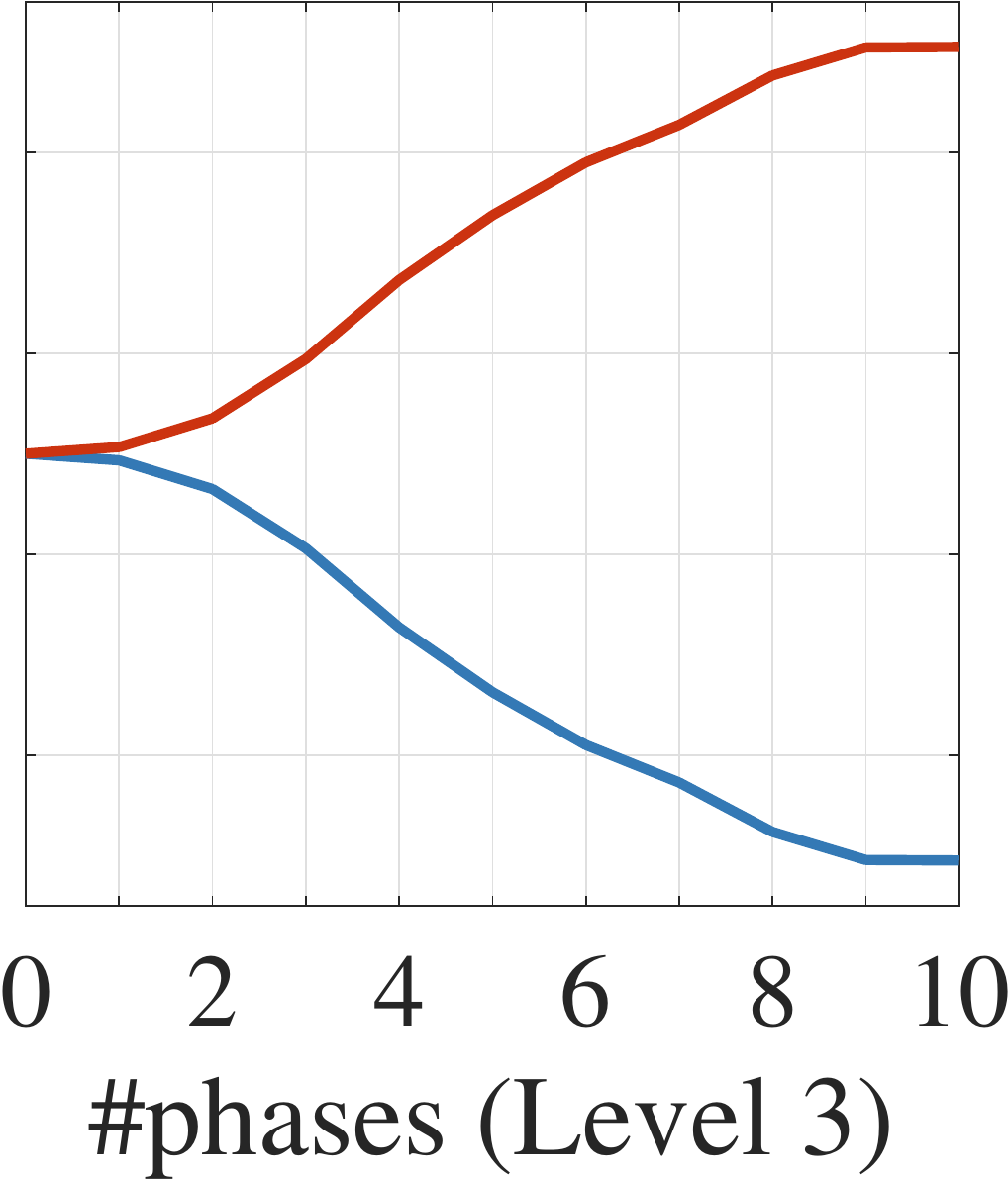}
}
\subfigure[ImageNet ($N$=10)]{
\newincludegraphics{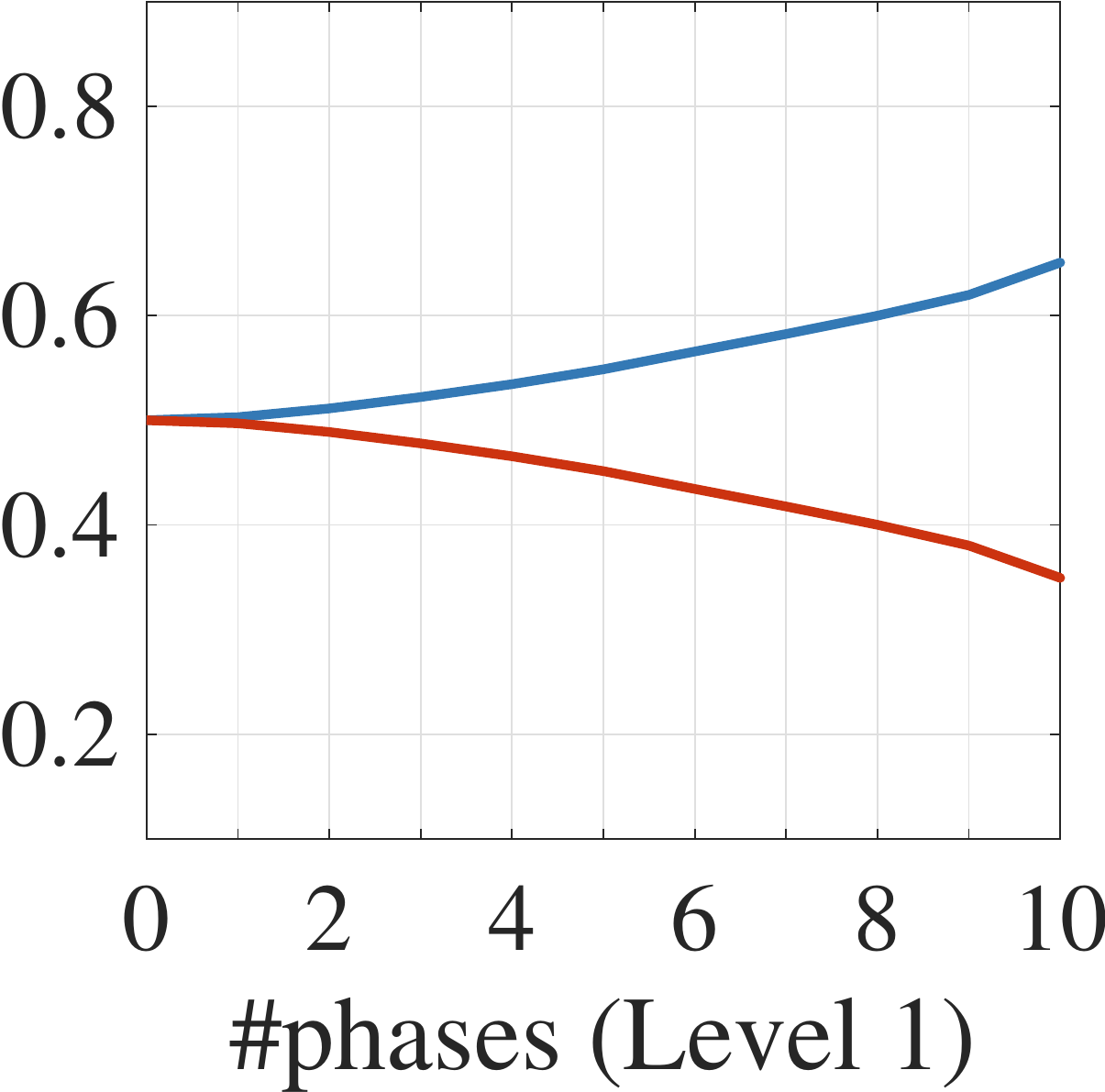}
\newincludegraphics{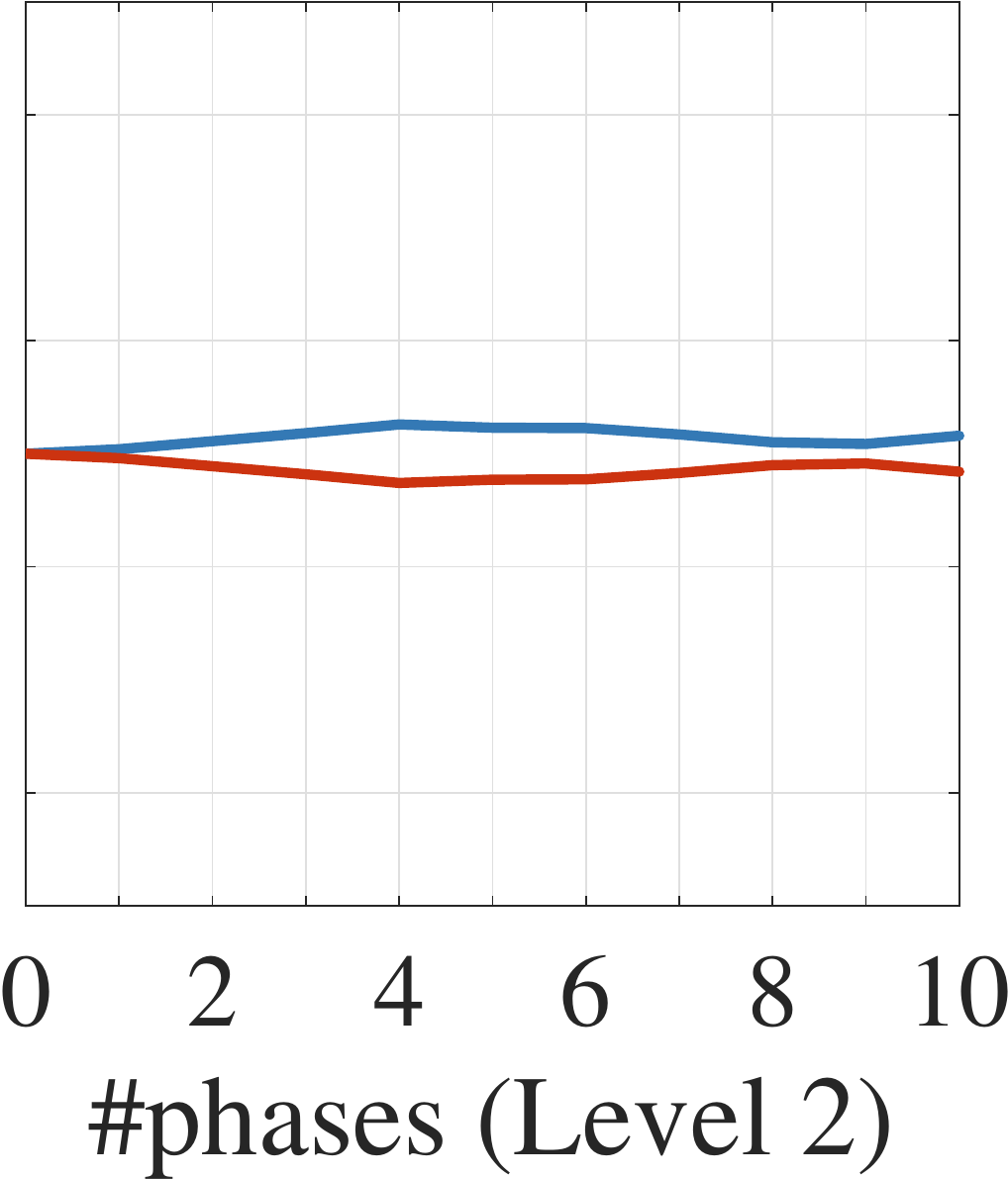}
\newincludegraphics{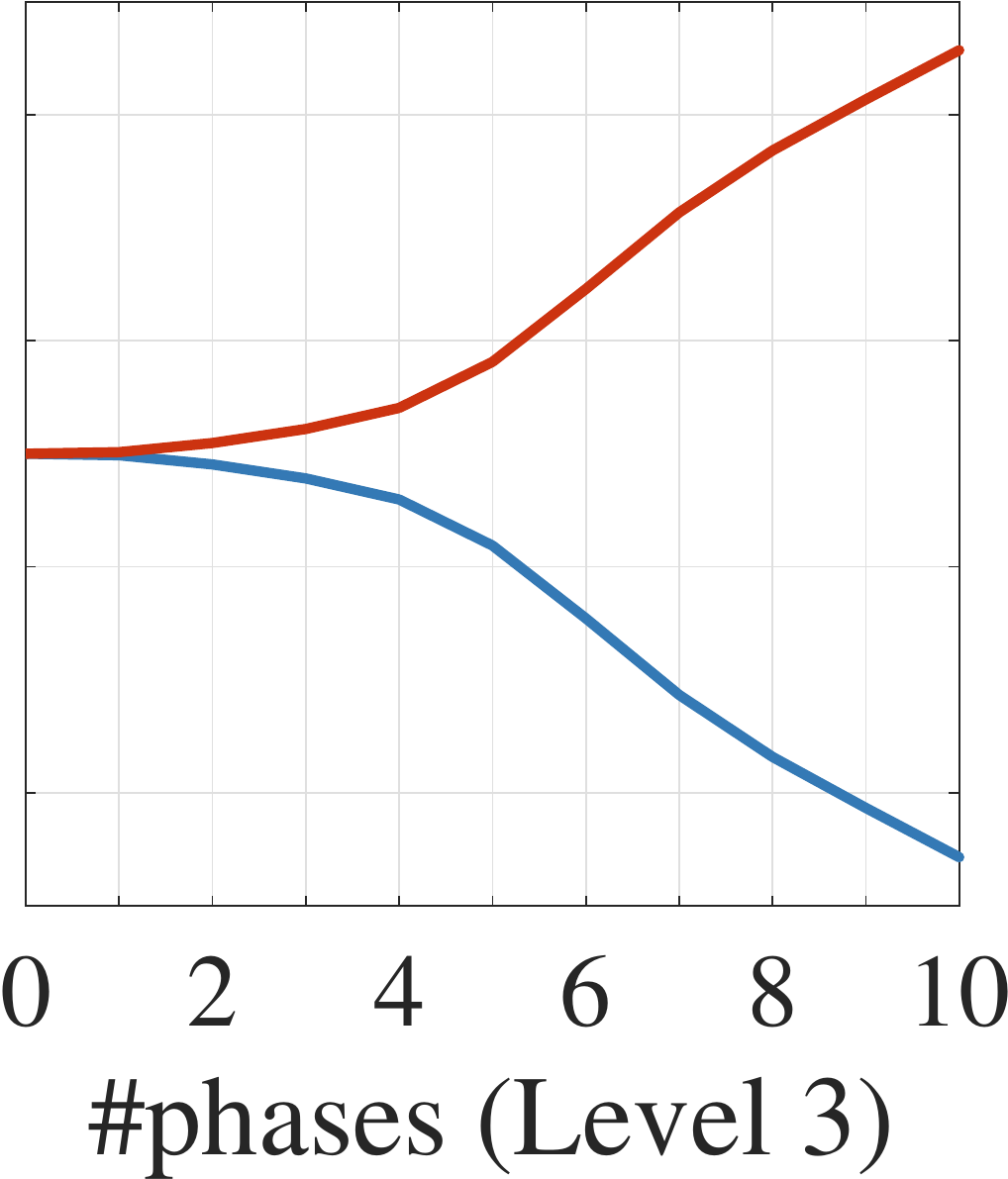}
}
\cotronlcaptionvsapce
\vspace{0.2cm}
\caption{The values of $\alpha_{\eta}$ and $\alpha_{\phi}$ adapted \emph{for each residual level} and \emph{in each incremental phase}.  All curves are
smoothed with a rate of $0.8$ for better visualization. 
}
\label{figure_values_plots}
\cotronlvsapce
\vspace{-0.5cm}
\end{figure}

%% file: sections/5_conclusions.tex
\section{Conclusions}
\label{sec_conclusion}

We introduce a novel network architecture AANets specially for CIL. 
Our main contribution lies in addressing the issue of stability-plasticity dilemma in CIL by a simple modification on plain ResNets --- applying two types of residual blocks to respectively and specifically learn stability and plasticity at each residual level, and then aggregating them as a final representation.
To achieve efficient aggregation, we adapt the level-specific and phase-specific weights in an end-to-end manner. 
Our overall approach is \emph{generic} and can be easily incorporated into existing CIL methods to boost their performance.

\vspace{0.1cm}
\myparagraph{Acknowledgments.} {This research is supported by A*STAR under its AME YIRG Grant (Project No. A20E6c0101), the Singapore Ministry of Education (MOE) Academic Research Fund (AcRF) Tier 1, Alibaba Innovative Research (AIR) programme, Major Scientific Research Project of Zhejiang Lab (No. 2019DB0ZX01), and Max Planck Institute for Informatics.}

%% file: supplementary/sections/6_supplementary.tex
\beginsupp
\setcounter{section}{0}
\renewcommand\thesection{\Alph{section}}
\noindent
{\Large {\textbf{Supplementary Materials}}}
\\

These supplementary materials include the results for different CIL settings(\S\ref{subsec_ablation_same_cls}), ``strict memory budget'' experiments (\S\ref{subsec_stric_memory}), additional ablation results (\S\ref{subsec_ablation}), additional plots (\S\ref{supp_sec_plots}), more visualization results 
(\S\ref{subsec_more_visualization}), and the execution steps of our source code with PyTorch (\S\ref{suppsec_code}).

\section{Results for Different CIL Settings.}
\label{subsec_ablation_same_cls}
We provide more results on the setting with the same number of classes at all phases~\cite{rajasegaran2020itaml} in the second block (\textbf{``same \# of cls''}) of Table~\ref{table_supp_same_cls}. For example, $N$=$25$ indicates $100$ classes evenly come in $25$ phases, so $4$ new classes arrive in each phase (including the $0$-th phase). 
Further, in this table, each entry represents an accuracy of the \textbf{last phase} (since all-phase accuracies are not comparable to our original setting) averaged over $3$ runs;
and ``update $\theta_{\mathrm{base}}$'' means that $\theta_{\mathrm{base}}$ is updated as $\theta_{\mathrm{base}}\gets\phi_i \odot \theta_{\mathrm{base}}$ after each phase. 
All results are under ``strict memory budget'' and ``all''+``scaling'' settings, so $\phi_i$ indicate the meta-learned weights of SS operators.
The results show that 
1)~``\emph{w/} AANets'' performs best in all settings and brings consistent improvements; and 2)~``update $\theta_{\mathrm{base}}$'' is helpful for CIFAR-100 but harmful for ImageNet-Subset.
\input{supplementary/tables/same_num_of_cls}

\section{Strict Memory Budget Experiments}
\label{subsec_stric_memory}
\input{supplementary/tables/sota_strict_memory} 

In Table~\ref{table_sota_stric_memory}, 
we present the results of $4$ state-of-the-art methods \emph{w/} and
\emph{w/o} AANetss as a plug-in architecture, under the ``strict memory budget'' setting which strictly controls the total memory shared by the exemplars and the model parameters.
For example, if we incorporate AANetss to LUCIR~\cite{hou2019lucir}, we need to reduce the number of exemplars to balance the additional memory introduced by AANetss (as AANetss take around $20\%$ more parameters than the plain ResNets used in LUCIR~\cite{hou2019lucir}).
As a result, we reduce the numbers of exemplars for AANetss from $20$ to $13$, $16$ and $19$, respectively, for CIFAR-100, ImageNet-Subset, and ImageNet, in the ``strict memory budget'' setting.
For CIFAR-100, we use $530k$ additional parameters, so we need to reduce $530k \mathrm{floats}\times4\mathrm{bytes/float}\div(32\times32\times3\mathrm{bytes}/\mathrm{image})\div100\mathrm{classes}\approx6.9\mathrm{images}/\mathrm{class}$, and $\lceil 6.9 \rceil=7\mathrm{images}/\mathrm{class}$.
For ImageNet-Subset, we use $12.6M$ additional parameters, so we need to reduce $12.6M \mathrm{floats}\times4\mathrm{bytes/float}\div(224\times224\times3\mathrm{bytes}/\mathrm{image})\div100\mathrm{classes}\approx3.3\mathrm{images}/\mathrm{class}$, and $\lceil 3.3 \rceil=4\mathrm{images}/\mathrm{class}$.
For ImageNet, we use $12.6M$ additional parameters, so we need to reduce $12.6M \mathrm{floats}\times4\mathrm{bytes/float}\div(224\times224\times3\mathrm{bytes}/\mathrm{image})\div100\mathrm{classes}\approx0.3\mathrm{images}/\mathrm{class}$, and $\lceil 0.3 \rceil=1\mathrm{image}/\mathrm{class}$. 
From Table~\ref{table_sota_stric_memory}, we can see that our approach of using AANetss still achieves the top performances in all CIL settings even if the ``strict memory budget'' is applied. 

\section{More Ablation Results}
\label{subsec_ablation}
\input{supplementary/tables/more_ablation}
In Table~\ref{table:ablation_add}, we supplement the ablation results obtained in more settings. ``$4\times$'' denotes that we use $4$ same-type blocks at each residual level. 
Comparing Row 7 to Row 2 (Row 5) shows the efficiency of using different types of blocks for representing stability and plasticity.

\section{Additional Plots}
\label{supp_sec_plots}

In Figures~\ref{figure_acc_plots}, we present the phase-wise accuracies obtained on CIFAR-100, ImageNet-Subset and ImageNet, respectively. ``Upper
Bound'' shows the results of joint training with all previous data accessible in every phase. We can observe that our method achieves the highest accuracies in almost every phase of different settings.
In Figures~\ref{figure_supp_values_cifar} and \ref{figure_supp_values_imgnetsub}, we supplement the plots for the values of $\alpha_{\eta}$ and $\alpha_{\phi}$ learned on the CIFAR-100 and ImageNet-Subset ($N$=$5$, $25$).
All curves are
smoothed with a rate of $0.8$ for a better visualization.

\section{More Visualization Results}
\label{subsec_more_visualization}
Figure~\ref{figure_gradcam_supp} below 
shows the activation maps of a ``goldfinch'' sample (seen in Phase 0) in different-phase models (ImageNet-Subset, $N$=$5$). 
Notice 
that the plastic block gradually loses its attention on this sample (i.e., forgets it), while the stable block retains it. AANets benefit from its stable blocks.
\input{supplementary/figures/more_visualization}

\section{Source Code in PyTorch}
\label{suppsec_code}

We provide our PyTorch code on  \href{https://class-il.mpi-inf.mpg.de/}{https://class-il.mpi-inf.mpg.de/}. 
To run this repository, we kindly advise you to install Python 3.6 and PyTorch 1.2.0 with Anaconda.

\input{supplementary/figures/acc_plot}

\input{supplementary/figures/3_value_plot_cifar}
\input{supplementary/figures/4_value_plot_imgnet}

%% file: supplementary/tables/same_num_of_cls.tex
\setlength{\tabcolsep}{0.9mm}{
\begin{table}[h]
  \vspace{-0.1cm}
  \small
  \centering
  \begin{tabular}{lcccccccc}
  \toprule
    % \\[-14pt]
  \multirow{2.5}{*}{Last-phase acc. (\%)} &  \multicolumn{3}{c}{\emph{CIFAR-100}} && \multicolumn{3}{c}{\emph{ImageNet-Subset}} \\
  \cmidrule{2-4} \cmidrule{6-8}
    % \\[-14pt]
   &  $N$=5 & 10  & 25 && 5 & 10 & 25  \\
    %  \\[-14pt]
   \midrule
    %  \\[-14pt]
    LUCIR (50 cls in Phase 0) &  54.3 & 50.3 & 48.4 && 60.0 & 57.1 & 49.3\\
     % \\[-14pt]
    \cellcolor{mygray-bg}{\ \ \emph{w/} AANets} &  \cellcolor{mygray-bg}{\textbf{58.6}} & \cellcolor{mygray-bg}{\textbf{56.7}} & \cellcolor{mygray-bg}{\textbf{53.3}} &\cellcolor{mygray-bg}{}& \cellcolor{mygray-bg}{64.3} & \cellcolor{mygray-bg}{58.0} & \cellcolor{mygray-bg}{56.5}\\
    %\\[-14pt]
   \midrule
    %  \\[-14pt]
    LUCIR (\textbf{same \# of cls}) &  52.1 & 44.9 & 40.6 && 60.3 & 52.5 & 53.3\\
     %  \\[-14pt]
     {\ \ \emph{w/} AANets, update $\theta{_\mathrm{base}}$} &  {54.3} & {47.4} & {42.4} && {61.4} & {52.5} & {48.2}\\ 
     %   \\[-14pt]
    \cellcolor{mygray-bg}{\ \ \emph{w/} AANets} &  \cellcolor{mygray-bg}{52.6} & \cellcolor{mygray-bg}{46.1} & \cellcolor{mygray-bg}{41.9} &\cellcolor{mygray-bg}{}& \cellcolor{mygray-bg}{\textbf{68.8}} & \cellcolor{mygray-bg}{\textbf{60.8}} & \cellcolor{mygray-bg}{\textbf{56.8}}\\
    %   \\[-14pt]
  \bottomrule
\end{tabular}
  \vspace{0.2cm}
  \caption{\mycaptionsupp{Supplementary to Table~\textcolor{red}{1}.} Last-phase accuracies (\%) for different class-incremental learning (CIL) settings. }
  \label{table_supp_same_cls}
  \vspace{-0.3cm}
\end{table}
}

%% file: supplementary/tables/sota_strict_memory.tex
\setlength{\tabcolsep}{1.0mm}{
\begin{table*}[ht]
  \small
  \centering
  \begin{tabular}{lccccccccccc}
  \toprule
   \multirow{2.5}{*}{Method} & \multicolumn{3}{c}{\emph{CIFAR-100}} && \multicolumn{3}{c}{\emph{ImageNet-Subset}} && \multicolumn{3}{c}{\emph{ImageNet}}\\
  \cmidrule{2-4} \cmidrule{6-8} \cmidrule{10-12}
   & $N$=5 & 10  & 25 && 5 & 10 & 25 && 5 & 10 & 25 \\
    \midrule
    iCaRL~\cite{rebuffi2017icarl} & 57.12\tiny{$\pm0.50$} & 52.66\tiny{$\pm0.89$} & 48.22\tiny{$\pm0.76$} && 65.44\tiny{$\pm0.35$} & 59.88\tiny{$\pm0.83$} & 52.97\tiny{$\pm1.02$} && 51.50\tiny{$\pm0.43$} & 46.89\tiny{$\pm0.35$} & 43.14\tiny{$\pm0.67$}  \\
   \cellcolor{mygray-bg}{ \ \ \emph{w/} {AANetss} (ours)}  & \cellcolor{mygray-bg}{63.91\tiny{$\pm0.52$}} & \cellcolor{mygray-bg}{{{57.65}}\tiny{$\pm0.81$}} &  \cellcolor{mygray-bg}{{{52.10}}\tiny{$\pm0.87$}} &\cellcolor{mygray-bg}{}& \cellcolor{mygray-bg}{{{71.37}}\tiny{$\pm0.57$}} &  \cellcolor{mygray-bg}{{{66.34}}\tiny{$\pm0.61$}} &  \cellcolor{mygray-bg}{{{61.87}}\tiny{$\pm1.01$}} &\cellcolor{mygray-bg}{}& \cellcolor{mygray-bg}{{{63.65}}\tiny{$\pm1.02$}} &  \cellcolor{mygray-bg}{{{61.14}}\tiny{$\pm0.59$}} &  \cellcolor{mygray-bg}{{{55.91}}\tiny{$\pm0.95$}} \\
    \cellcolor{mygray-bg}{} & \cellcolor{mygray-bg}{\redtext{64.22}}\tiny{\redtext{$\pm0.42$}} &  \cellcolor{mygray-bg}{\redtext{60.26}}\tiny{\redtext{$\pm0.73$}} &  \cellcolor{mygray-bg}{\redtext{56.43}}\tiny{\redtext{$\pm0.81$}} &\cellcolor{mygray-bg}{}&  \cellcolor{mygray-bg}{\redtext{73.45}}\tiny{\redtext{$\pm0.51$}} &  \cellcolor{mygray-bg}{\redtext{71.78}}\tiny{\redtext{$\pm0.64$}} &  \cellcolor{mygray-bg}{\redtext{69.22}}\tiny{\redtext{$\pm0.83$}} &\cellcolor{mygray-bg}{}&  \cellcolor{mygray-bg}{\redtext{63.91}}\tiny{\redtext{$\pm0.59$}} &  \cellcolor{mygray-bg}{\redtext{61.28}}\tiny{\redtext{$\pm0.49$}} &  \cellcolor{mygray-bg}{\redtext{56.97}}\tiny{\redtext{$\pm0.86$}} \\
    \midrule
    LUCIR~\cite{hou2019lucir}  & 63.17\tiny{$\pm0.87$} & 60.14\tiny{$\pm0.73$} & 57.54\tiny{$\pm0.43$} && 70.84\tiny{$\pm0.69$} & 68.32\tiny{$\pm0.81$} & 61.44\tiny{$\pm0.91$} && 64.45\tiny{$\pm0.32$} & 61.57\tiny{$\pm0.23$} & 56.56\tiny{$\pm0.36$} \\
   \cellcolor{mygray-bg}{ \ \ \emph{w/} {AANetss} (ours)}  & \cellcolor{mygray-bg}{66.46\tiny{$\pm0.45$} } & \cellcolor{mygray-bg}{{{65.38}}\tiny{$\pm0.53$}} &  \cellcolor{mygray-bg}{{{61.79}}\tiny{$\pm0.51$}} &\cellcolor{mygray-bg}{}& \cellcolor{mygray-bg}{{{72.21}}\tiny{$\pm0.87$}} &  \cellcolor{mygray-bg}{{{69.10}}\tiny{$\pm0.90$}} &  \cellcolor{mygray-bg}{{{67.10}}\tiny{$\pm0.54$}} &\cellcolor{mygray-bg}{}& \cellcolor{mygray-bg}{{{64.83}}\tiny{$\pm0.50$}} &  \cellcolor{mygray-bg}{{{62.34}}\tiny{$\pm0.65$}} &  \cellcolor{mygray-bg}{{{60.49}}\tiny{$\pm0.78$}} \\
  \cellcolor{mygray-bg}{ }  & \cellcolor{mygray-bg}{\redtext{66.74}}\tiny{\redtext{$\pm0.37$}} & \cellcolor{mygray-bg}{\redtext{65.29}}\tiny{\redtext{$\pm0.43$}} & \cellcolor{mygray-bg}{\redtext{63.50}}\tiny{\redtext{$\pm0.61$}} &\cellcolor{mygray-bg}{}&  \cellcolor{mygray-bg}{\redtext{72.55}}\tiny{\redtext{$\pm0.67$}} &  \cellcolor{mygray-bg}{\redtext{69.22}}\tiny{\redtext{$\pm0.72$}} &  \cellcolor{mygray-bg}{\redtext{67.60}}\tiny{\redtext{$\pm0.39$}} &\cellcolor{mygray-bg}{}&  \cellcolor{mygray-bg}{\redtext{64.94}}\tiny{\redtext{$\pm0.25$}} &  \cellcolor{mygray-bg}{\redtext{62.39}}\tiny{\redtext{$\pm0.61$}} &  \cellcolor{mygray-bg}{\redtext{60.68}}\tiny{\redtext{$\pm0.58$}} \\
    \midrule
    Mnemonics~\cite{liu2020mnemonics} & {{63.34}}\tiny{$\pm0.62$} & {{62.28}}\tiny{$\pm0.43$} &  {{60.96}}\tiny{$\pm0.72$} &&  {{72.58}}\tiny{$\pm0.85$} &  {{71.37}}\tiny{$\pm0.56$} &  {{69.74}\tiny{$\pm0.39$}} &&  {{64.54}}\tiny{$\pm0.49$} &  {{63.01}}\tiny{$\pm0.57$} &  {{61.00}}\tiny{$\pm0.71$} \\
   \cellcolor{mygray-bg}{ \ \ \emph{w/} {AANetss} (ours)}  & \cellcolor{mygray-bg}{66.12\tiny{$\pm0.00$}} & \cellcolor{mygray-bg}{{{65.10}}\tiny{$\pm0.00$}} &  \cellcolor{mygray-bg}{{{61.83}}\tiny{$\pm0.00$}} &\cellcolor{mygray-bg}{}& \cellcolor{mygray-bg}{{{72.88}}\tiny{$\pm0.00$}} &  \cellcolor{mygray-bg}{{{71.50}}\tiny{$\pm0.00$}} &  \cellcolor{mygray-bg}{{{70.49}}\tiny{$\pm0.00$}} &\cellcolor{mygray-bg}{}& \cellcolor{mygray-bg}{{{65.21}}\tiny{$\pm0.76$}} &  \cellcolor{mygray-bg}{{{63.36}}\tiny{$\pm0.67$}} &  \cellcolor{mygray-bg}{{{61.37}}\tiny{$\pm0.80$}} \\
    \cellcolor{mygray-bg}{ } & \cellcolor{mygray-bg}{{\redtext{67.59}}}\tiny{\redtext{$\pm0.34$}} & \cellcolor{mygray-bg}{{\redtext{65.66}}}\tiny{\redtext{$\pm0.61$}} &  \cellcolor{mygray-bg}{\redtext{63.35}}\tiny{\redtext{$\pm0.72$}} &\cellcolor{mygray-bg}{}&  \cellcolor{mygray-bg}{\redtext{72.91}}\tiny{\redtext{$\pm0.53$}} &  \cellcolor{mygray-bg}{\redtext{71.93}}\tiny{\redtext{$\pm0.37$}} &  \cellcolor{mygray-bg}{\redtext{70.70}}\tiny{\redtext{$\pm0.45$}} &\cellcolor{mygray-bg}{}&  \cellcolor{mygray-bg}{{\redtext{65.23}}}\tiny{\redtext{$\pm0.62$}} &  \cellcolor{mygray-bg}{{\redtext{63.60}}}\tiny{\redtext{$\pm0.71$}} &  \cellcolor{mygray-bg}{\redtext{61.53}}\tiny{\redtext{$\pm0.29$}} \\
    \midrule
    PODNet-CNN~\cite{douillard2020podnet} & 64.83\tiny{$\pm1.11$} & 63.19\tiny{$\pm1.31$} &  60.72\tiny{$\pm1.54$} &&  75.54\tiny{$\pm0.29$} &  74.33\tiny{$\pm1.05$} &  68.31\tiny{$\pm2.77$} &&  66.95 &  64.13 & {{59.17}} \\
   \cellcolor{mygray-bg}{ \ \ \emph{w/} {AANetss} (ours)}  & \cellcolor{mygray-bg}{66.36\tiny{$\pm1.02$}} & \cellcolor{mygray-bg}{{{64.31}}\tiny{$\pm1.13$}} &  \cellcolor{mygray-bg}{{{61.80}}\tiny{$\pm1.24$}} &\cellcolor{mygray-bg}{}& \cellcolor{mygray-bg}{{{76.63}}\tiny{$\pm0.35$}} &  \cellcolor{mygray-bg}{{{75.00}}\tiny{$\pm0.78$}} &  \cellcolor{mygray-bg}{{{71.43}}\tiny{$\pm1.51$}} &\cellcolor{mygray-bg}{}& \cellcolor{mygray-bg}{{{67.80}}\tiny{$\pm0.87$}} &  \cellcolor{mygray-bg}{{{64.80}}\tiny{$\pm0.60$}} &  \cellcolor{mygray-bg}{{{61.01}}\tiny{$\pm0.97$}} \\
   \cellcolor{mygray-bg}{ }  & \cellcolor{mygray-bg}{\redtext{66.31}}\tiny{\redtext{$\pm0.87$}} & \cellcolor{mygray-bg}{\redtext{64.31}}\tiny{\redtext{$\pm0.90$}} &  \cellcolor{mygray-bg}{\redtext{62.31}}\tiny{\redtext{$\pm1.02$}} &\cellcolor{mygray-bg}{}& \cellcolor{mygray-bg}{ \redtext{76.96}}\tiny{\redtext{$\pm0.53$}} &  \cellcolor{mygray-bg}{\redtext{75.58}}\tiny{\redtext{$\pm0.74$}} &  \cellcolor{mygray-bg}{\redtext{71.78}}\tiny{\redtext{$\pm0.81$}} &\cellcolor{mygray-bg}{}& \cellcolor{mygray-bg}{ {\redtext{67.73}}\tiny{\redtext{$\pm0.71$}}} &  \cellcolor{mygray-bg}{{\redtext{64.85}}}\tiny{\redtext{$\pm0.53$}} &  \cellcolor{mygray-bg}{\redtext{{61.78}}\tiny{\redtext{$\pm0.61$}}} \\
  \bottomrule

\end{tabular}
%\cotronlcaptionvsapce
\vspace{0.2cm}
  \caption{\mycaptionsupp{Supplementary to Table~\textcolor{red}{2}.}
  \textbf{Using ``strict memory budget'' setting.}
  Average incremental accuracies (\%) of four state-of-the-art methods \emph{w/} and \emph{w/o} our AANetss as a plug-in architecture. The \redt{red lines} are the corresponding results in Table~\textcolor{red}{2} of the main paper.
}
  \label{table_sota_stric_memory}
  \cotronlvsapce
\end{table*}
}

%% file: supplementary/tables/more_ablation.tex
\setlength{\tabcolsep}{1.10mm}{
\begin{table*}[htp]
  \small
  \centering
  \vspace{-0.15cm}
	  \begin{tabular}{cccccccccccccccccccc}
      \toprule
       \multirow{2.5}{*}{Row} && \multicolumn{3}{c}{\multirow{2.5}{*}{Ablation Setting}}  && \multicolumn{6}{c}{\emph{CIFAR-100} (acc.\%)} && \multicolumn{6}{c}{\emph{ImageNet-Subset} (acc.\%)}\\
       \cmidrule{7-12} \cmidrule{14-19} 
       && &   &  &&Memory&FLOPs& \#Param & $N$=5 & 10  & 25  && Memory&FLOPs&\#Param & $N$=5 & 10  & 25\\
       \midrule
        1 & \multicolumn{4}{c}{single-branch ``all''~\cite{hou2019lucir}} &&7.64MB&70M&469K& 63.17 & 60.14 & 57.54 &&330MB&1.82G&11.2M& 70.84 & 68.32 & 61.44 \\
        2  & \multicolumn{4}{c}{``all'' + ``all''} &&9.43MB&140M&938K&  64.49 & 61.89 & 58.87 &&372MB&3.64G&22.4M& 69.72 & 66.69 & 63.29\\
        3  & \multicolumn{4}{c}{$4\times$ ``all''} &&13.01MB&280M&1.9M&  65.13 & 64.08 & 59.40 &&456MB&7.28G&44.8M& 70.12 & 67.31 & 64.00\\
      \midrule
        4 & \multicolumn{4}{c}{single-branch ``scaling''} &&7.64MB&70M&60K& 62.48 & 61.53 & 60.17 &&334MB&1.82G&1.4M& 71.29 & 68.88 & 66.75 \\
        5  & \multicolumn{4}{c}{``scaling'' + ``scaling''} &&9.43MB&140M&120K&  {65.13} & 64.08 & 62.50 &&382MB&3.64G&2.8M& 71.71 & 71.07 & 66.69\\
        6  & \multicolumn{4}{c}{$4\times$ ``scaling''} &&13.01MB&240M&280K& 66.00 & 64.67 & 63.16 &&478MB&3.64G&5.6M& 72.01 & 71.23 & 67.12\\
      \midrule
        7 & \multicolumn{4}{c}{``all'' + ``scaling''} &&9.66MB&140M&530K&  \highestablation{66.74} & \highestablation{65.29}
        & \highestablation{63.50} &&378MB&3.64G&12.6M& {{72.55}} &  {{69.22}} &  67.60 \\
        8 & \multicolumn{4}{c}{``all'' + ``frozen''} &&9.43MB&140M&469K&  65.62 & 64.05 & {63.67} &&372MB&3.64G&11.2M& 71.71 & 69.87 & 67.92\\
        9 & \multicolumn{4}{c}{``scaling'' + ``frozen''}  &&9.66MB&140M&60K&  64.71 & 63.65 & 62.89 &&378MB&3.64G&1.4M& \highestablation{73.01} & \highestablation{71.65} & \highestablation{70.30}\\
      \bottomrule
    \end{tabular}
    \centering
    \cotronlcaptionvsapce
    \vspace{0.2cm}
	\caption{\mycaptionsupp{Supplementary to Table~\textcolor{red}{1}.}
	More ablation study. ``$4\times$'' denotes that we use $4$ same-type blocks at each residual level.
	}
	\label{table:ablation_add}
	\cotronlvsapce
	\vspace{0.2cm}
    \end{table*}
    }

%% file: supplementary/figures/more_visualization.tex
\begin{figure}[h]
    \newcommand{\newincludegraphics}[1]{\includegraphics[height=0.46in]{#1}}
    \centering
    \vspace{0.0cm}
    \newincludegraphics{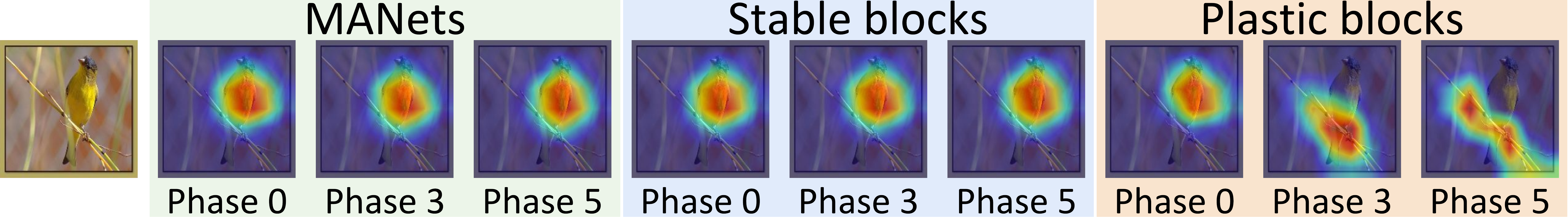}
    \vspace{-0.2cm}
    \caption{\mycaptionsupp{Supplementary to Figure~\textcolor{red}{3}.} The activation maps of a ``goldfinch'' sample (seen in Phase 0) in different-phase models (ImageNet-Subset; $N$=$5$). }
    \label{figure_gradcam_supp}
    \vspace{-0.4cm}
    \end{figure}

%% file: supplementary/figures/acc_plot.tex
\begin{figure*}
\newcommand{\newincludegraphics}[1]{\includegraphics[height=1.26in]{#1}}
\centering
\includegraphics[height=0.19in]{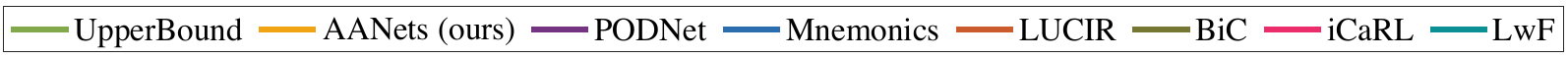}
\vspace{0.21cm}
\subfigure[CIFAR-100 ($100$ classes). In the $0$-th phase, $\theta_{\mathrm{base}}$ is trained on $50$ classes, the remaining  classes are given evenly in the subsequent phases.]{
\newincludegraphics{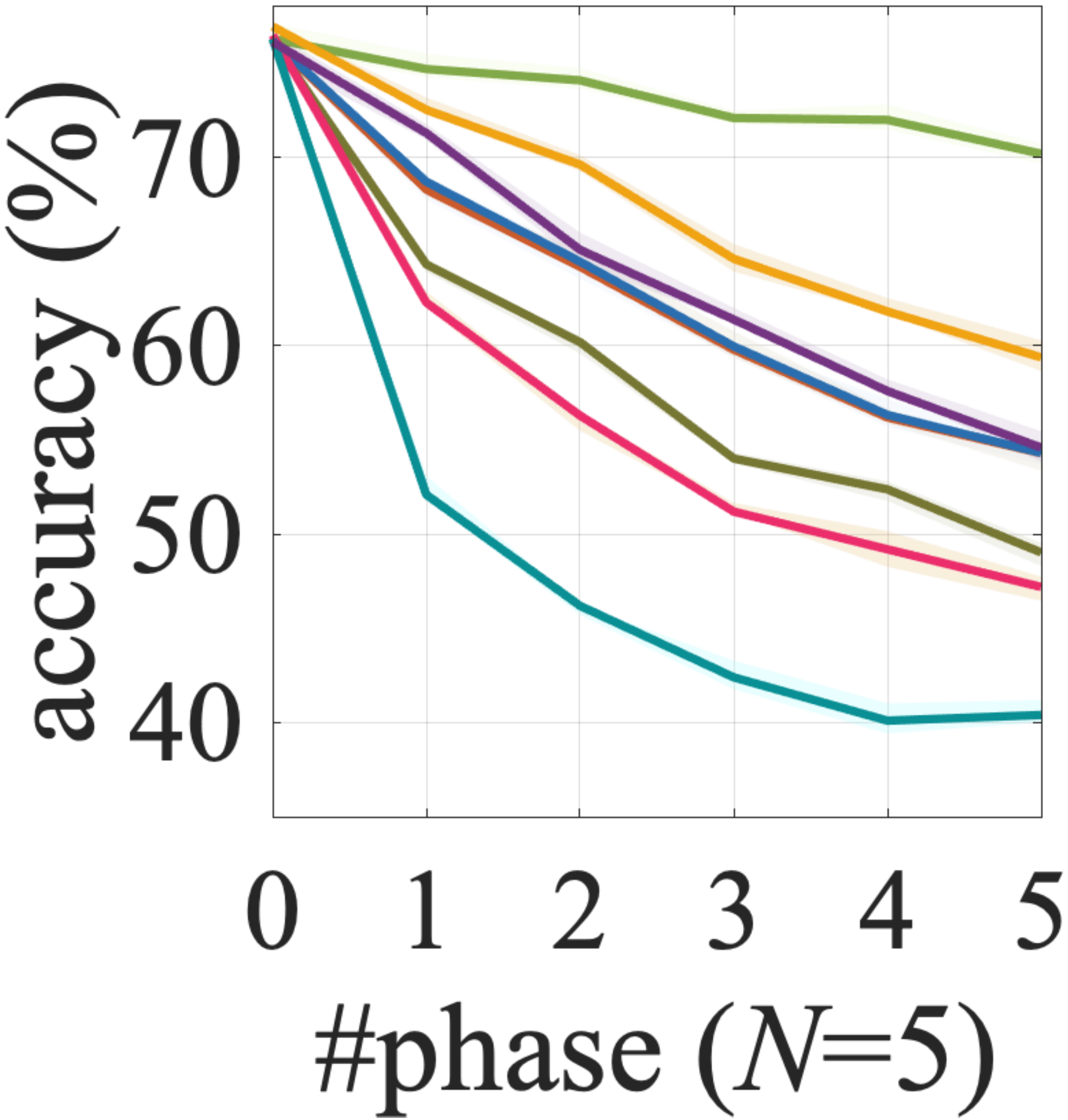}

\hspace{1mm}
\newincludegraphics{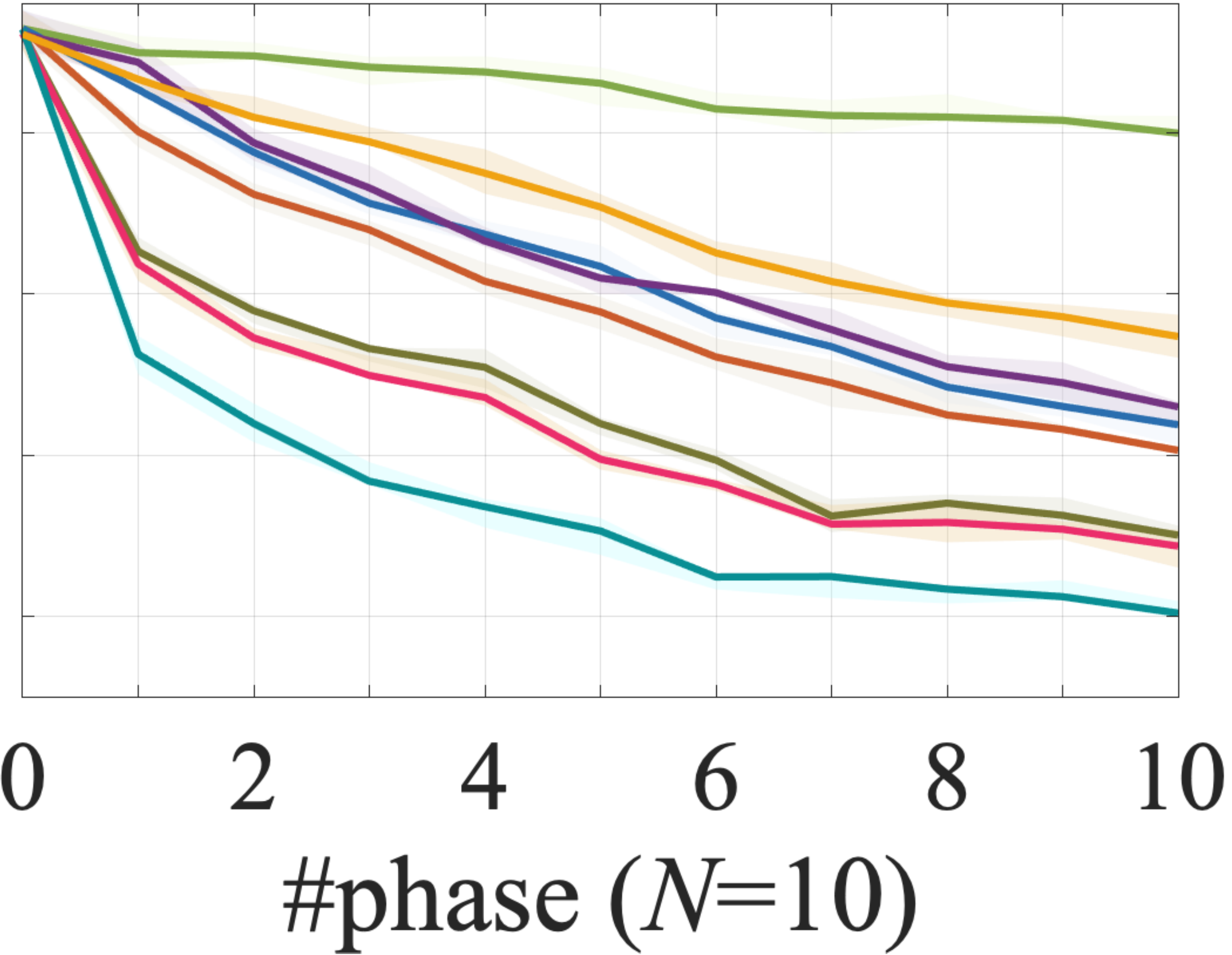}
\hspace{1mm}
\newincludegraphics{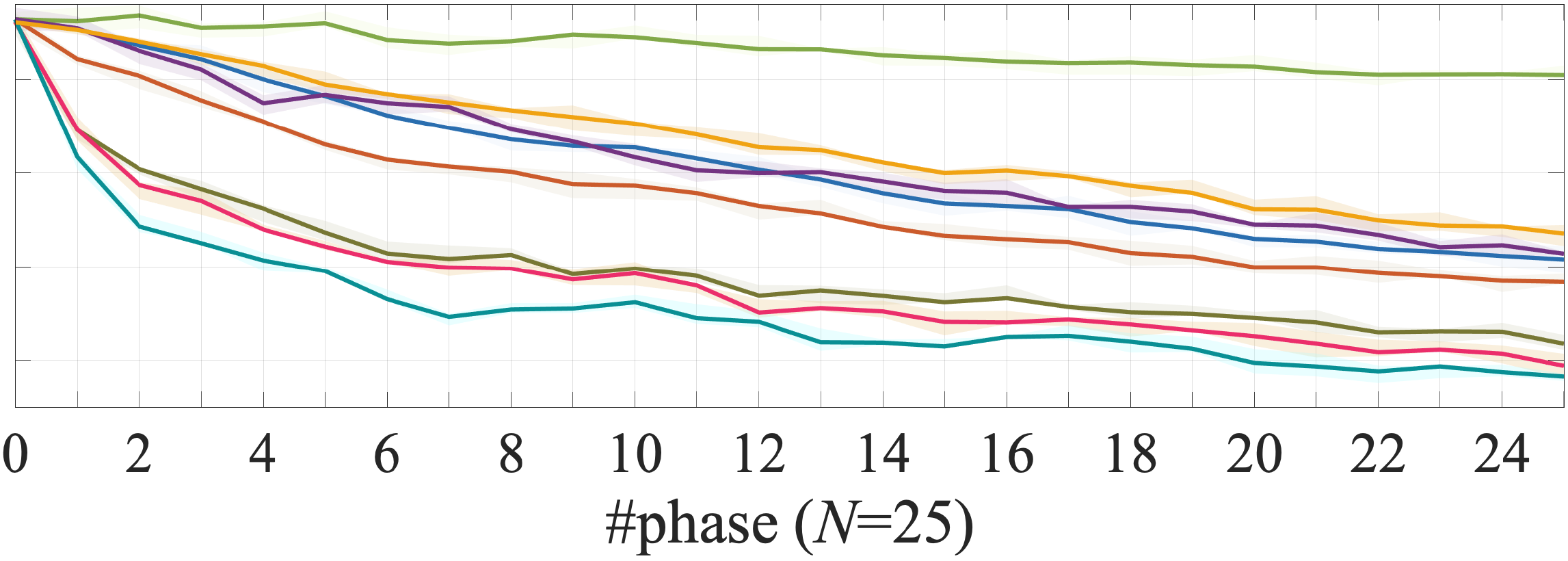}
}

\subfigure[ImageNet-Subset ($100$ classes). In the $0$-th phase, $\theta_{\mathrm{base}}$ is trained on $50$ classes, the remaining classes are given evenly in the subsequent phases.]{
\newincludegraphics{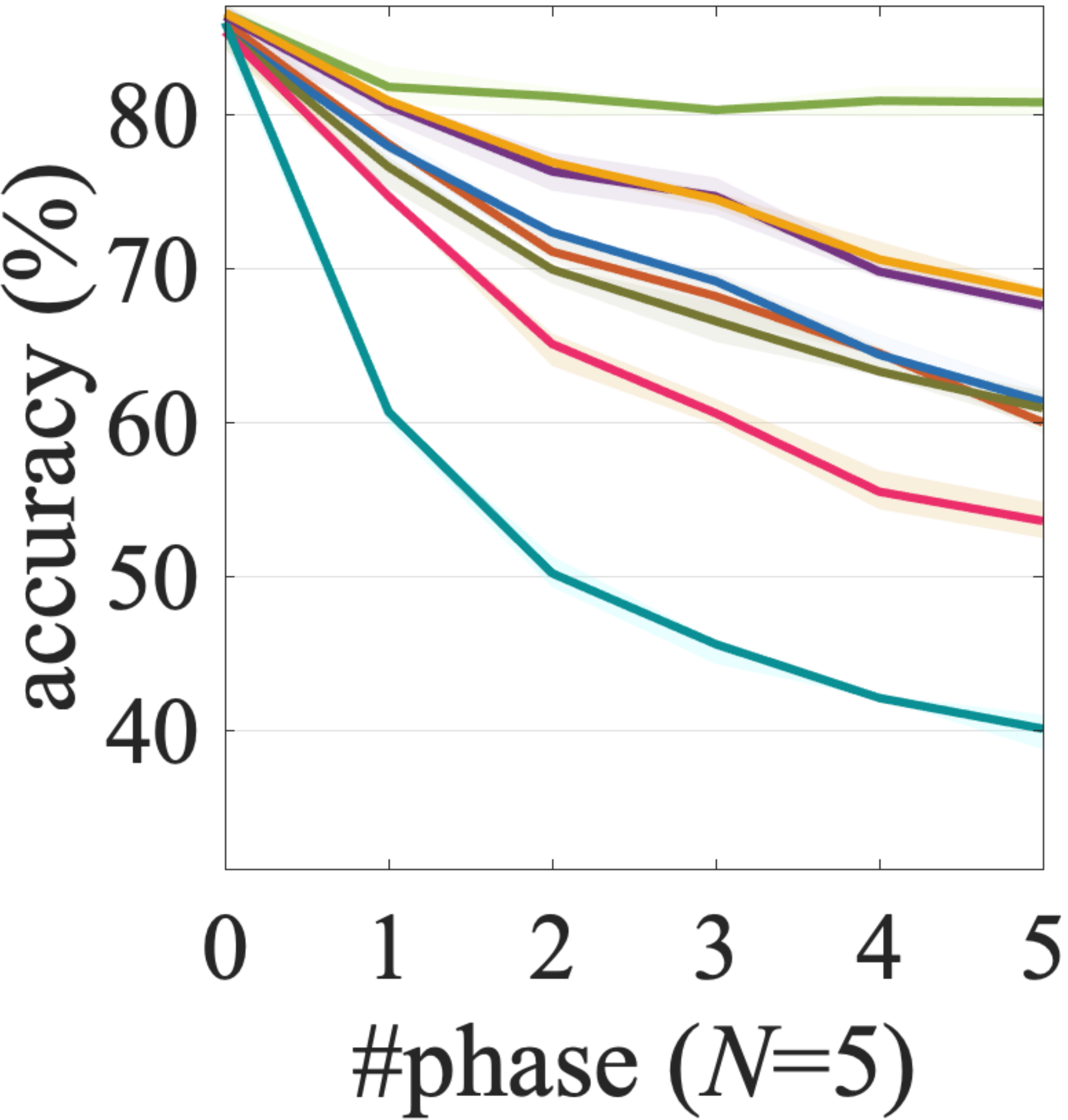}

\hspace{1mm}
\newincludegraphics{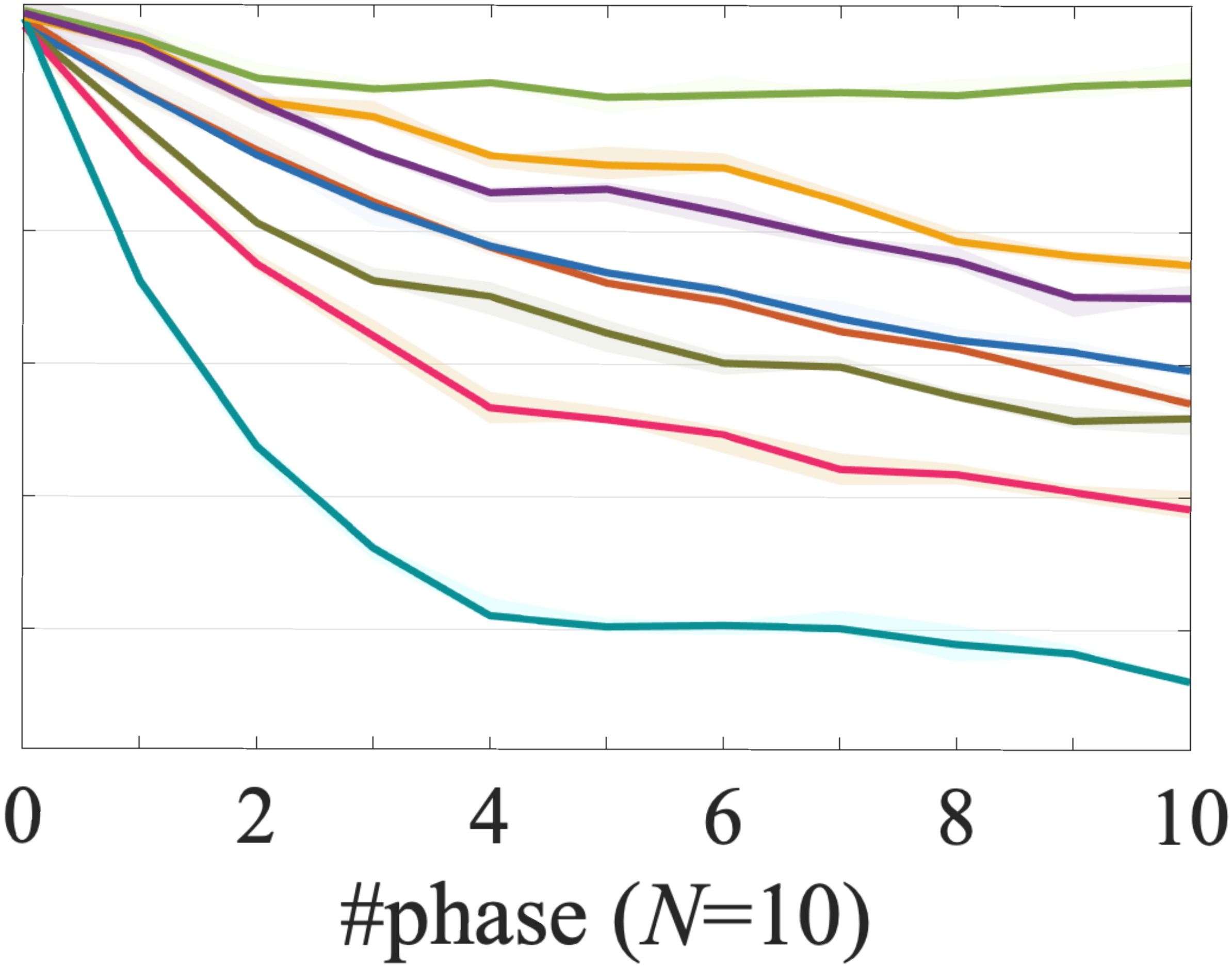}

\hspace{1mm}
\newincludegraphics{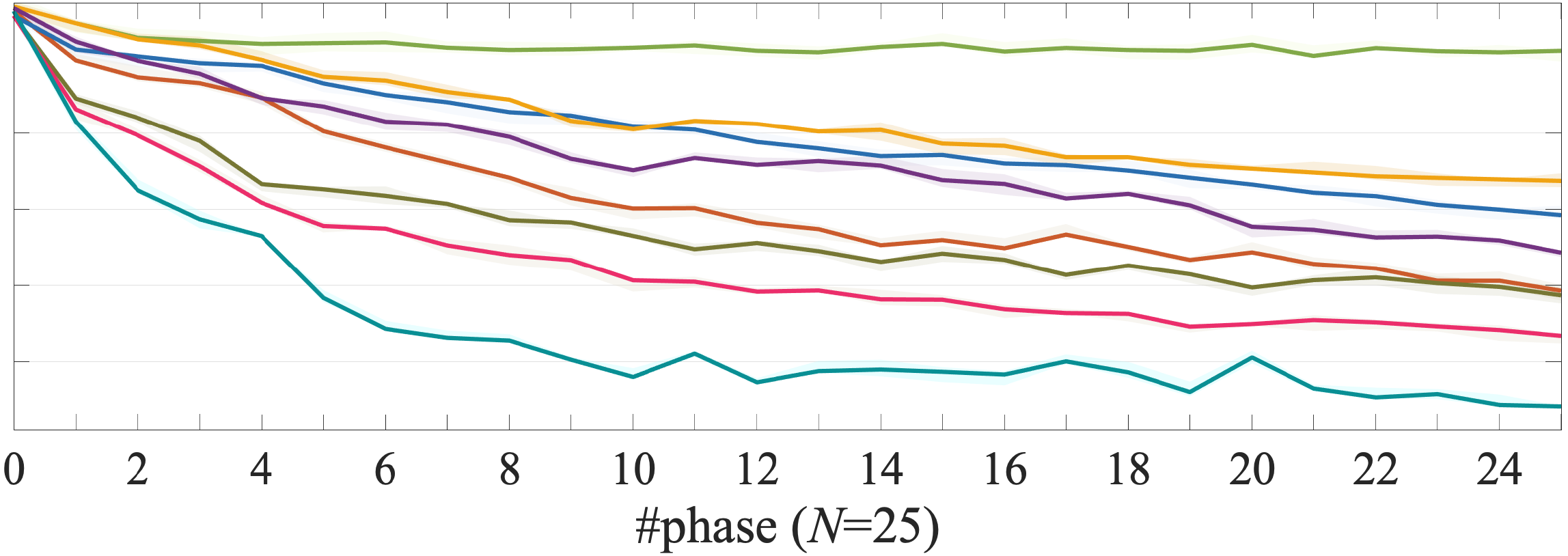}
}

\subfigure[ImageNet ($1000$ classes). In the $0$-th phase, $\theta_{\mathrm{base}}$ on is trained on $500$ classes, the remaining classes are given evenly in the subsequent phases.]{
\newincludegraphics{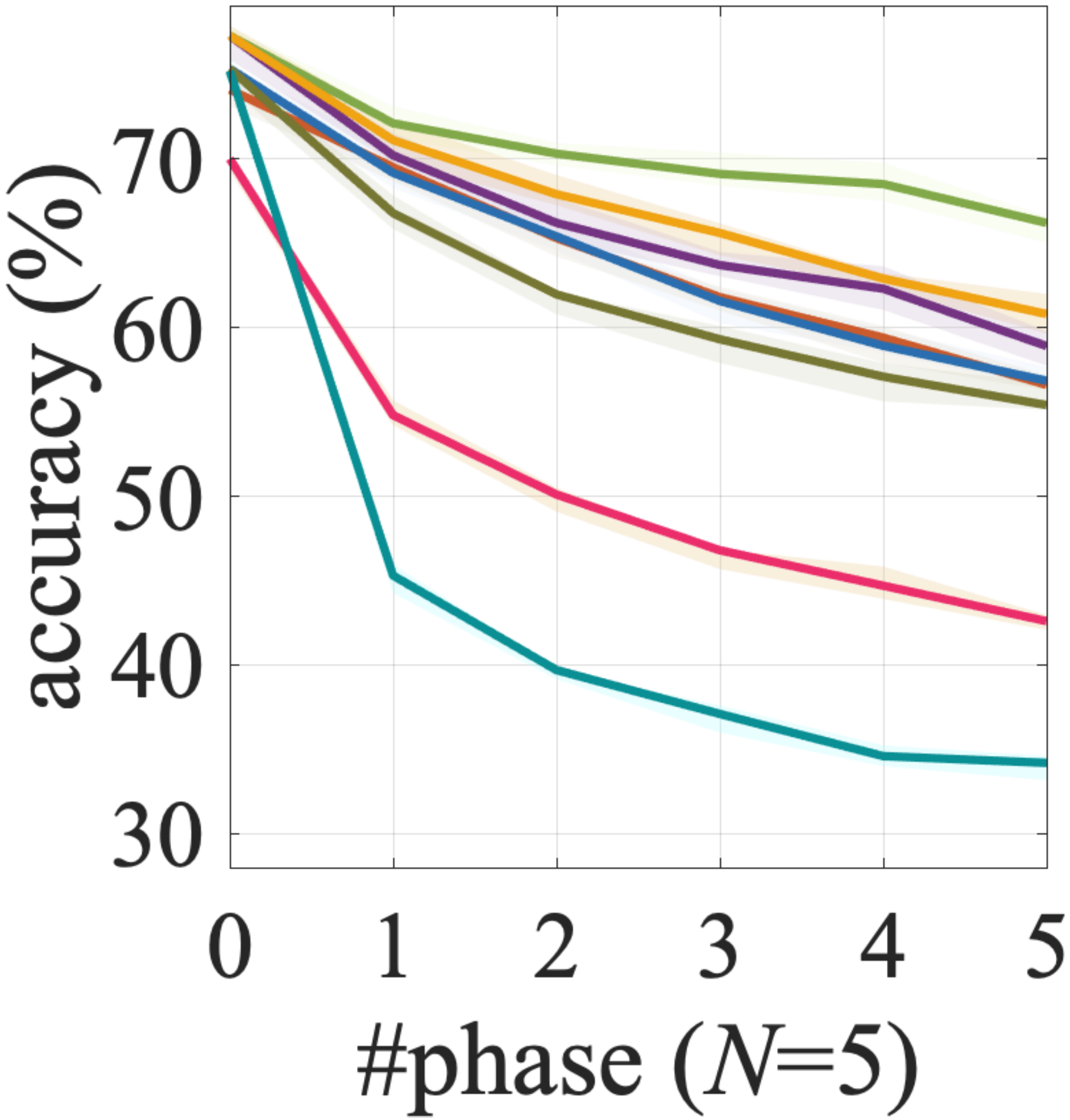}

\hspace{1mm}

\newincludegraphics{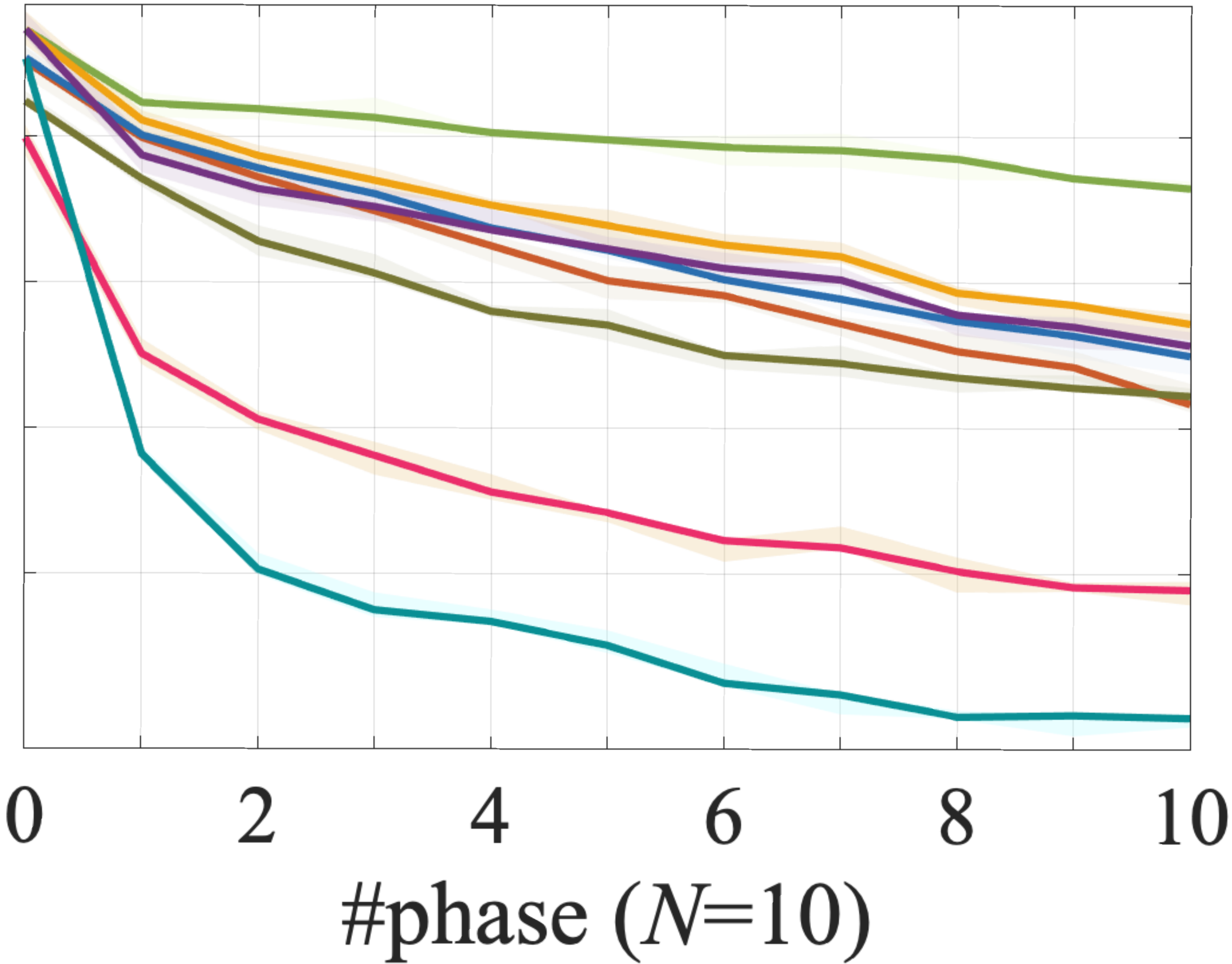}

\hspace{1mm}

\newincludegraphics{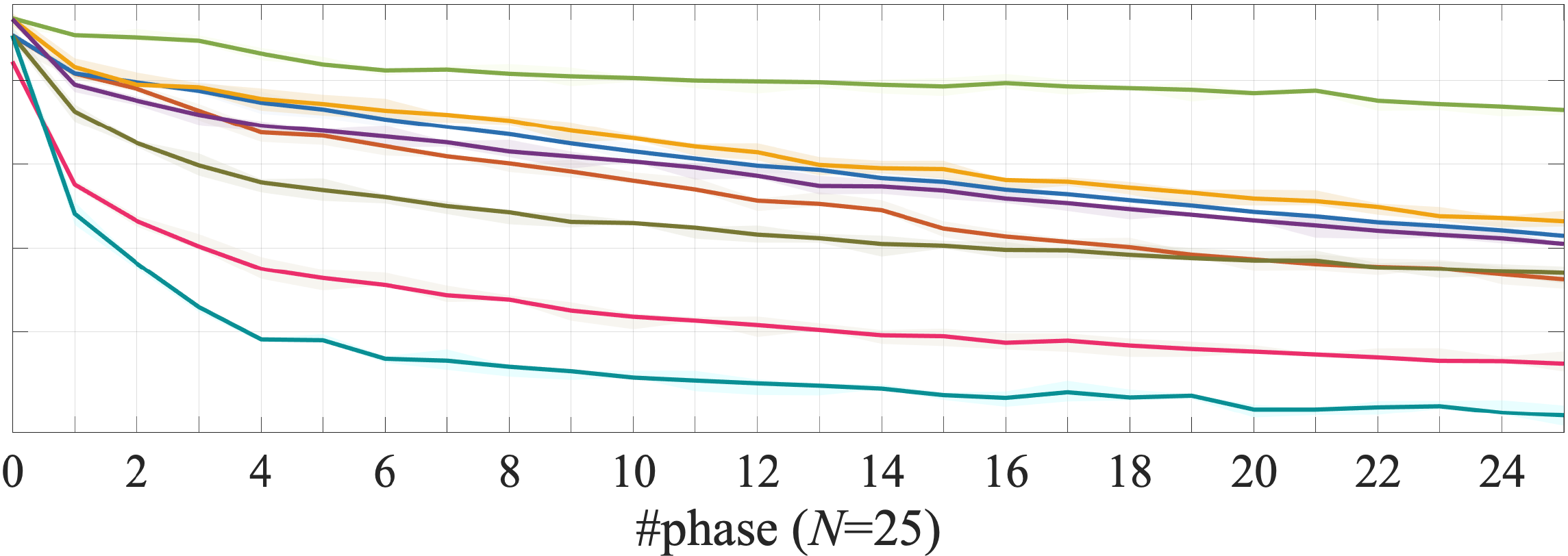}
}
\vspace{0.2cm}
\caption{\mycaptionsupp{Supplementary to Table~\textcolor{red}{2}.}Phase-wise accuracies ($\%$). 
Light-color ribbons are visualized to show the $95\%$ confidence intervals. 
Comparing methods: Upper Bound (the results of joint training with all previous data accessible in each phase); PODNet (2020)~\cite{douillard2020podnet}; Mnemonics (2020)~\cite{liu2020mnemonics}; LUCIR (2019)~\cite{hou2019lucir}; BiC (2019)~\cite{Wu2019LargeScale}; iCaRL (2017)~\cite{rebuffi2017icarl}; and LwF (2016)~\cite{Li18LWF}. 
}
\label{figure_acc_plots}
\cotronlvsapce
\end{figure*}

%% file: supplementary/figures/3_value_plot_cifar.tex
\begin{figure*}[ht]
\newcommand{\cifarincludegraphics}[1]{\includegraphics[height=1.5in]{#1}}
\centering
\subfigure[CIFAR-100, $N$=5]{
\cifarincludegraphics{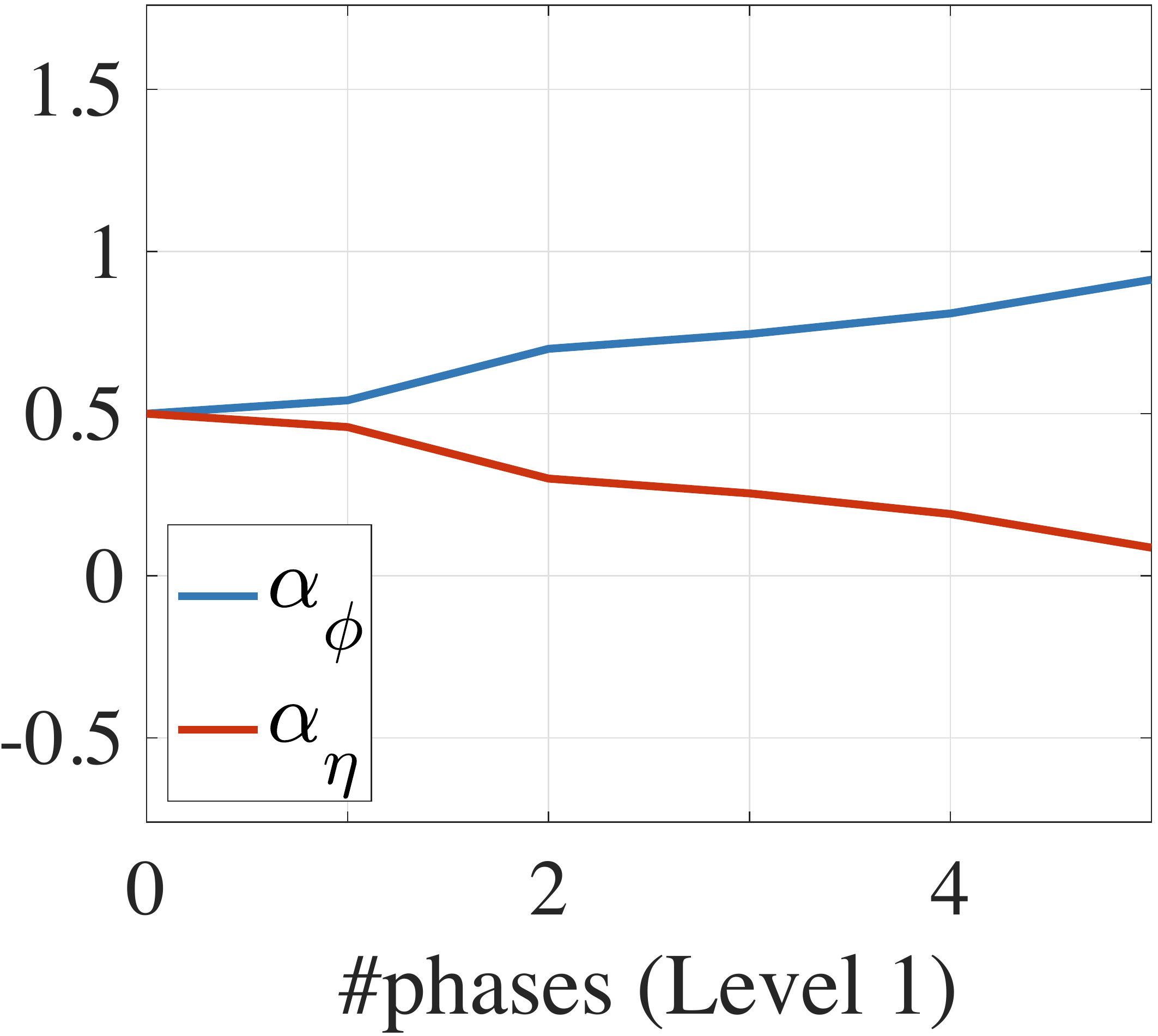}
\hspace{0.2cm}
\cifarincludegraphics{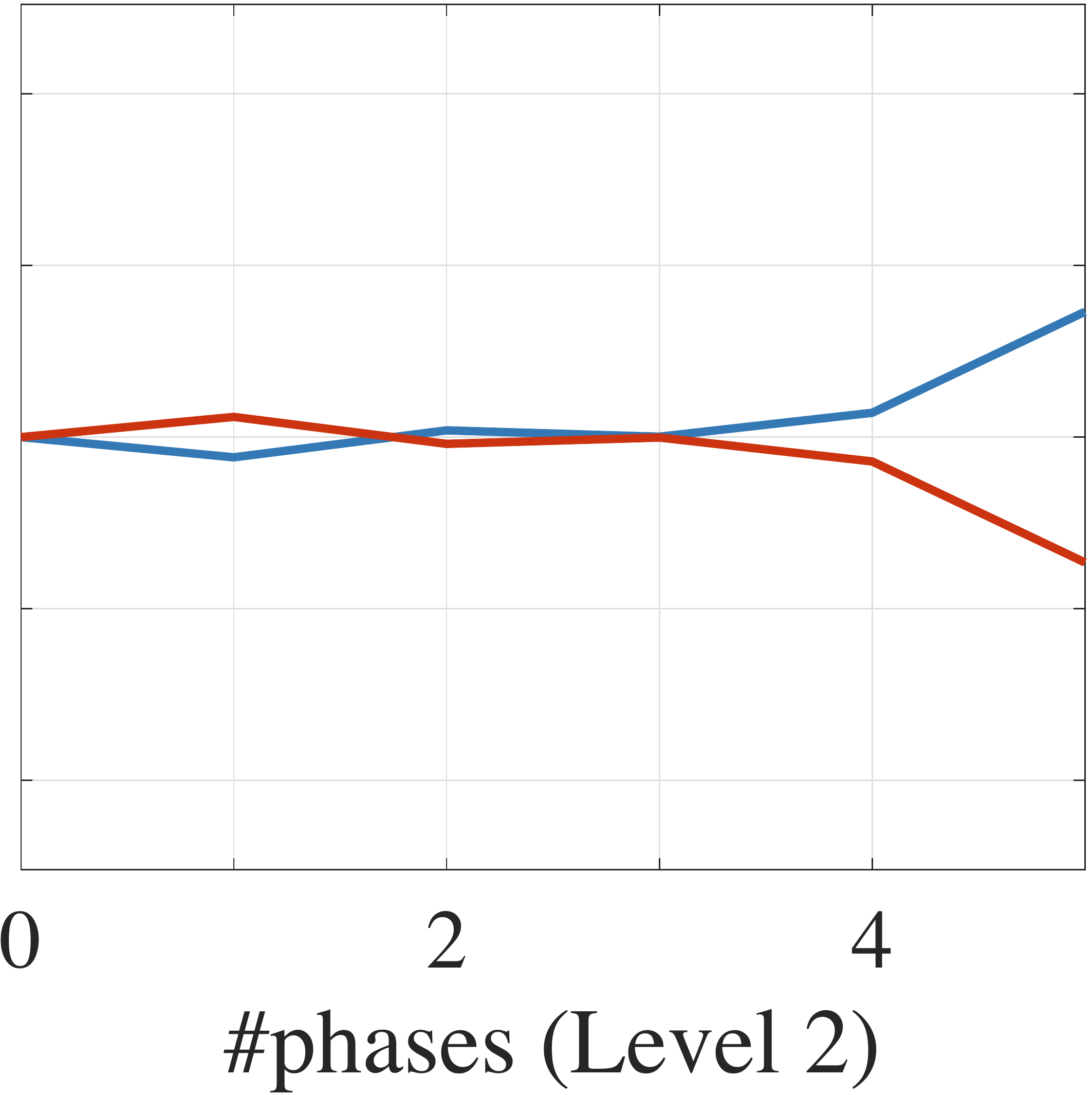}
\hspace{0.2cm}
\cifarincludegraphics{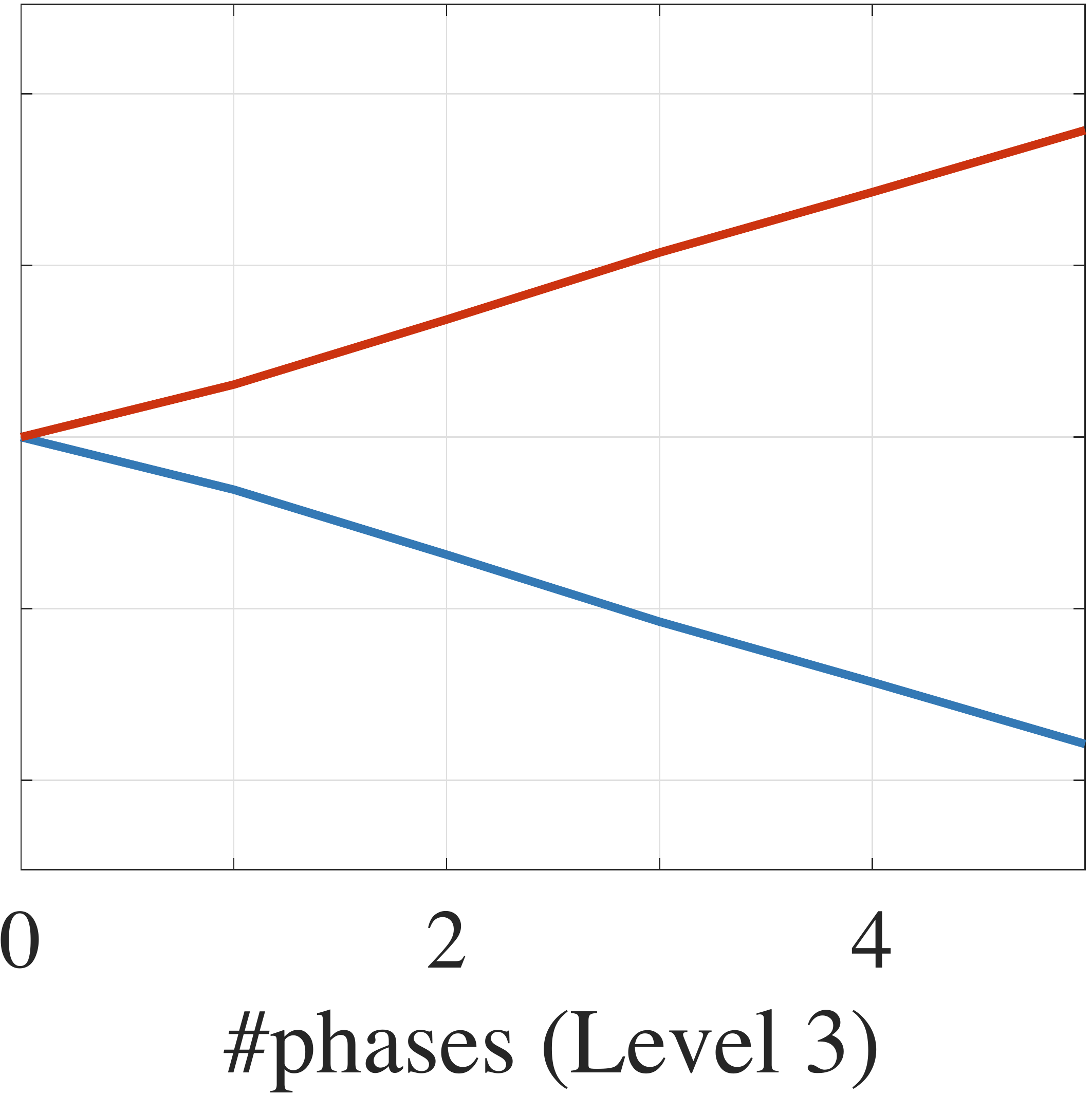}
}
\subfigure[CIFAR-100, $N$=25]{
\hspace{0.1cm}
\cifarincludegraphics{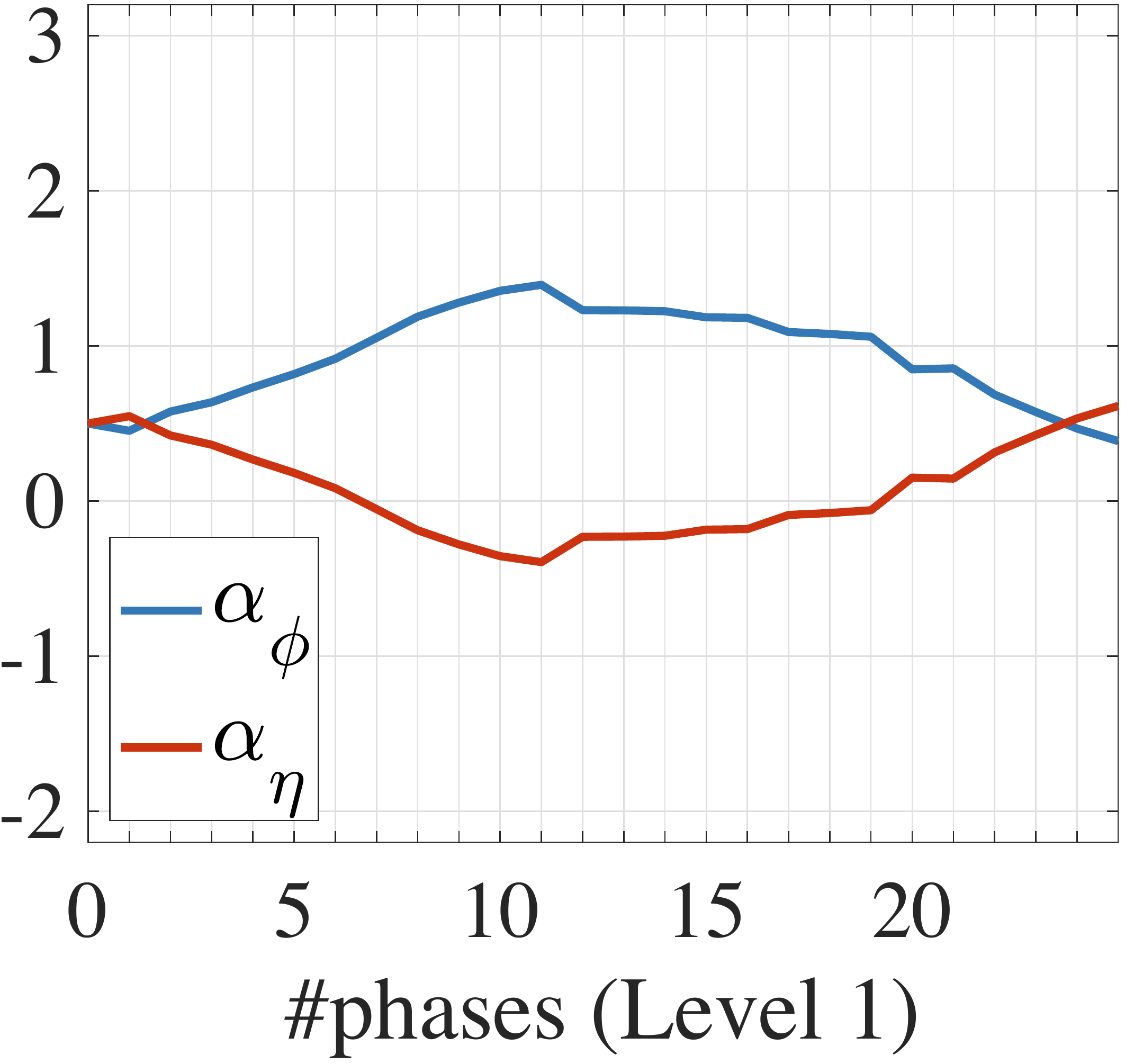}
\hspace{0.2cm}
\cifarincludegraphics{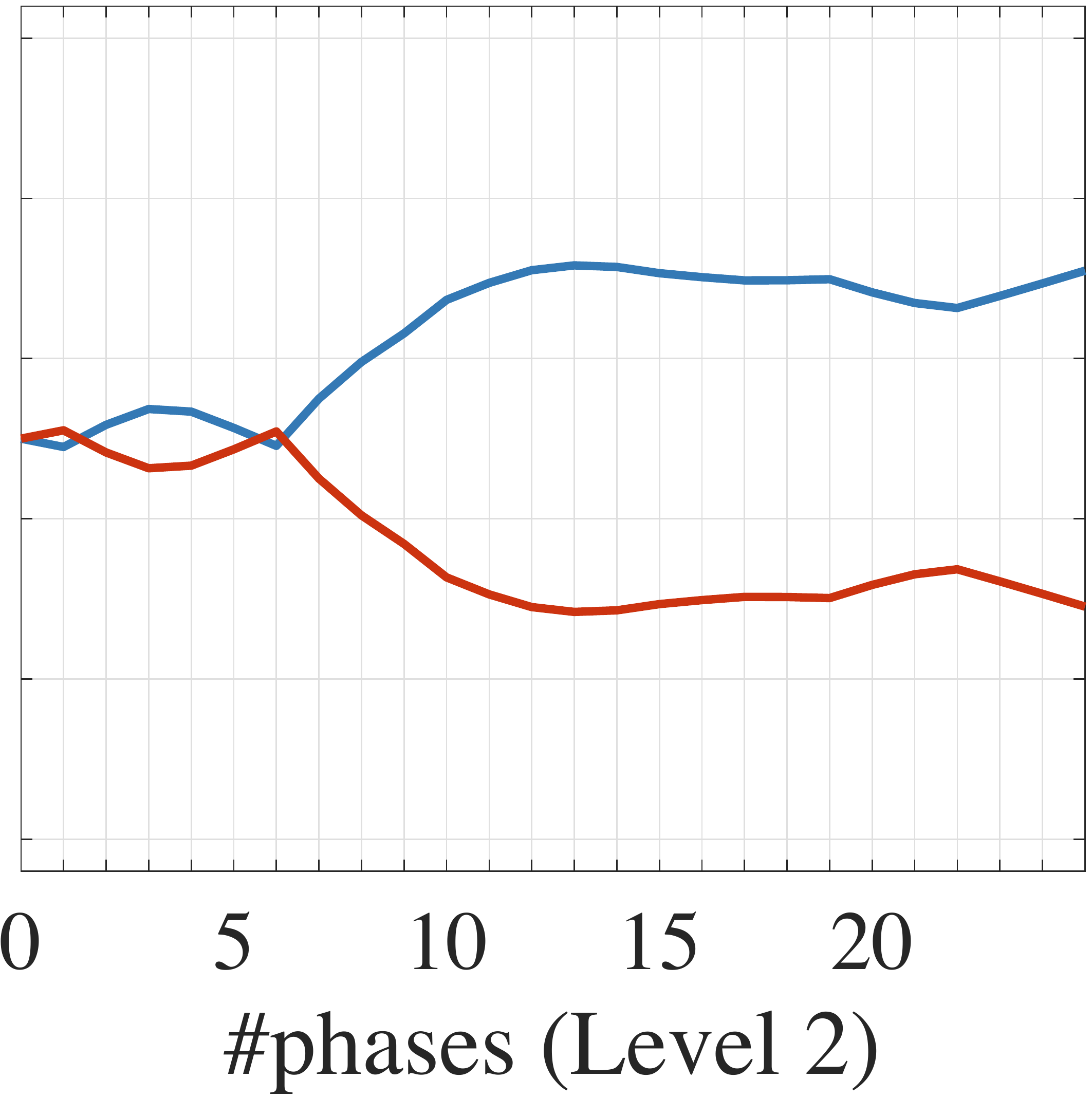}
\hspace{0.2cm}
\cifarincludegraphics{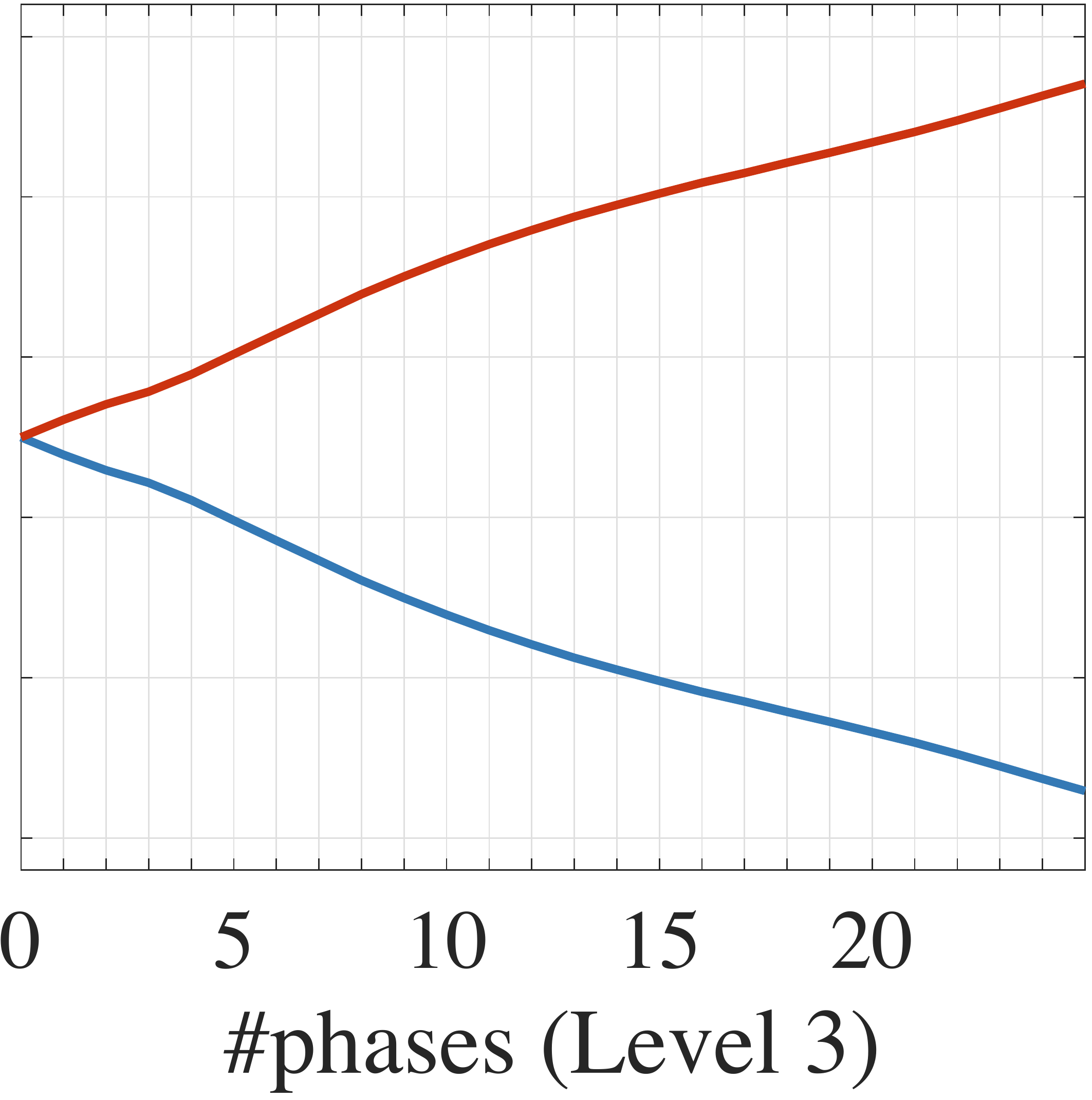}
}

\caption{\mycaptionsupp{Supplementary to Figure~\textcolor{red}{4}.} The changes of values for $\alpha_{\eta}$ and $\alpha_{\phi}$ on CIFAR-100.}
\label{figure_supp_values_cifar}
\end{figure*}

%% file: supplementary/figures/4_value_plot_imgnet.tex
\begin{figure*}[ht]
\newcommand{\cifarincludegraphics}[1]{\includegraphics[height=1.5in]{#1}}
\centering
\subfigure[ImageNet-Subset, $N$=5]{
\cifarincludegraphics{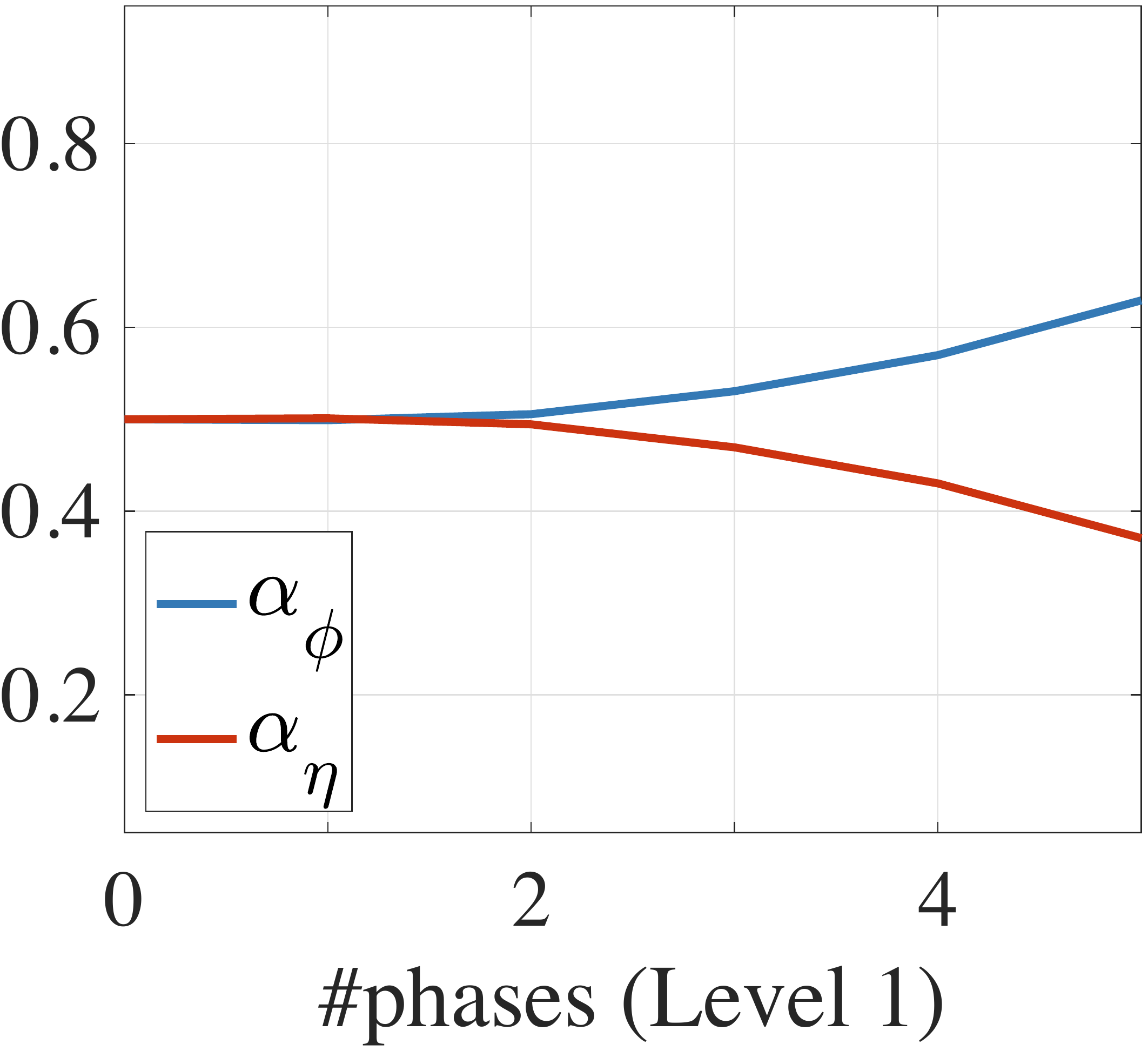}
\hspace{0.2cm}
\cifarincludegraphics{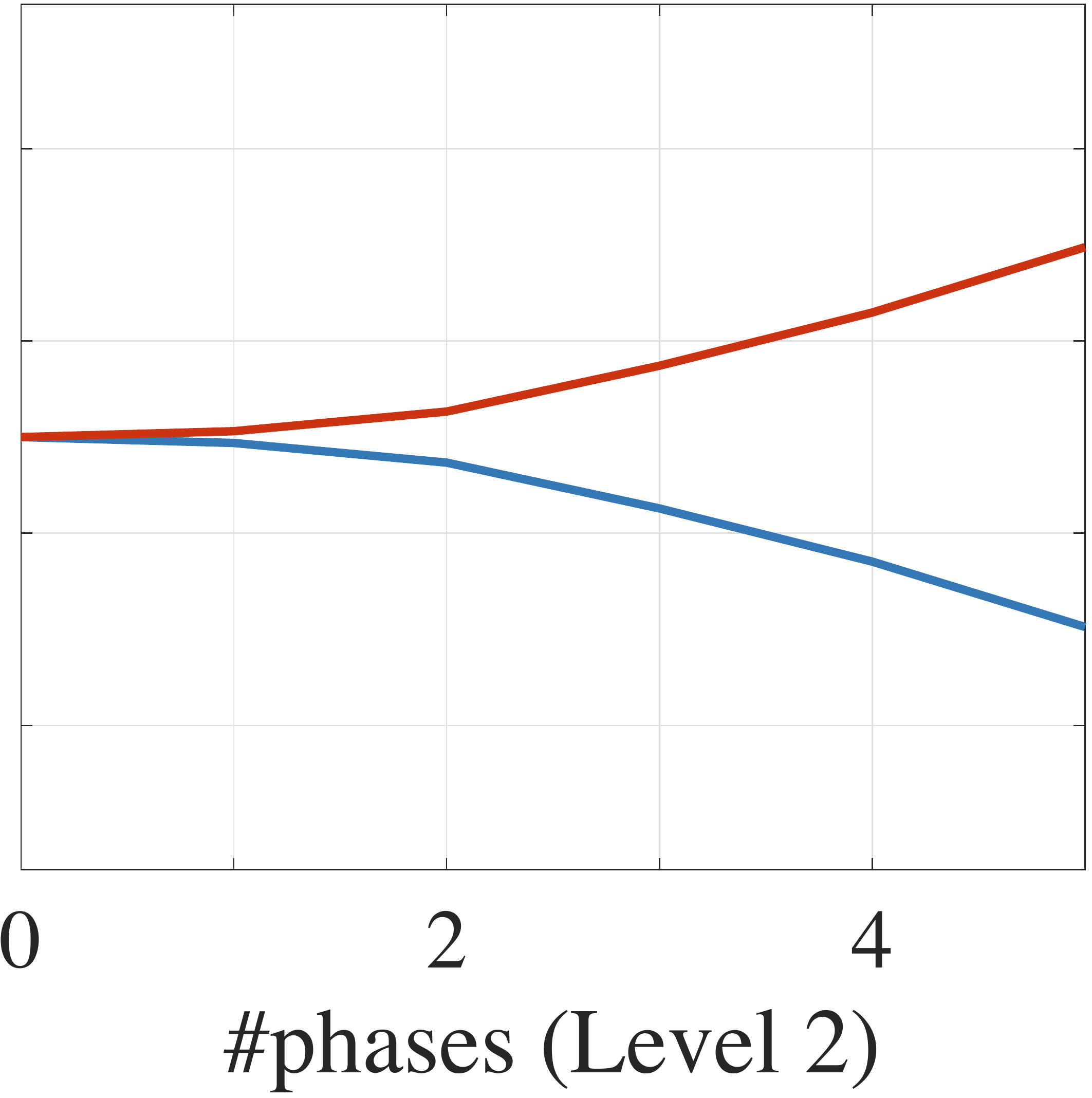}
\hspace{0.2cm}
\cifarincludegraphics{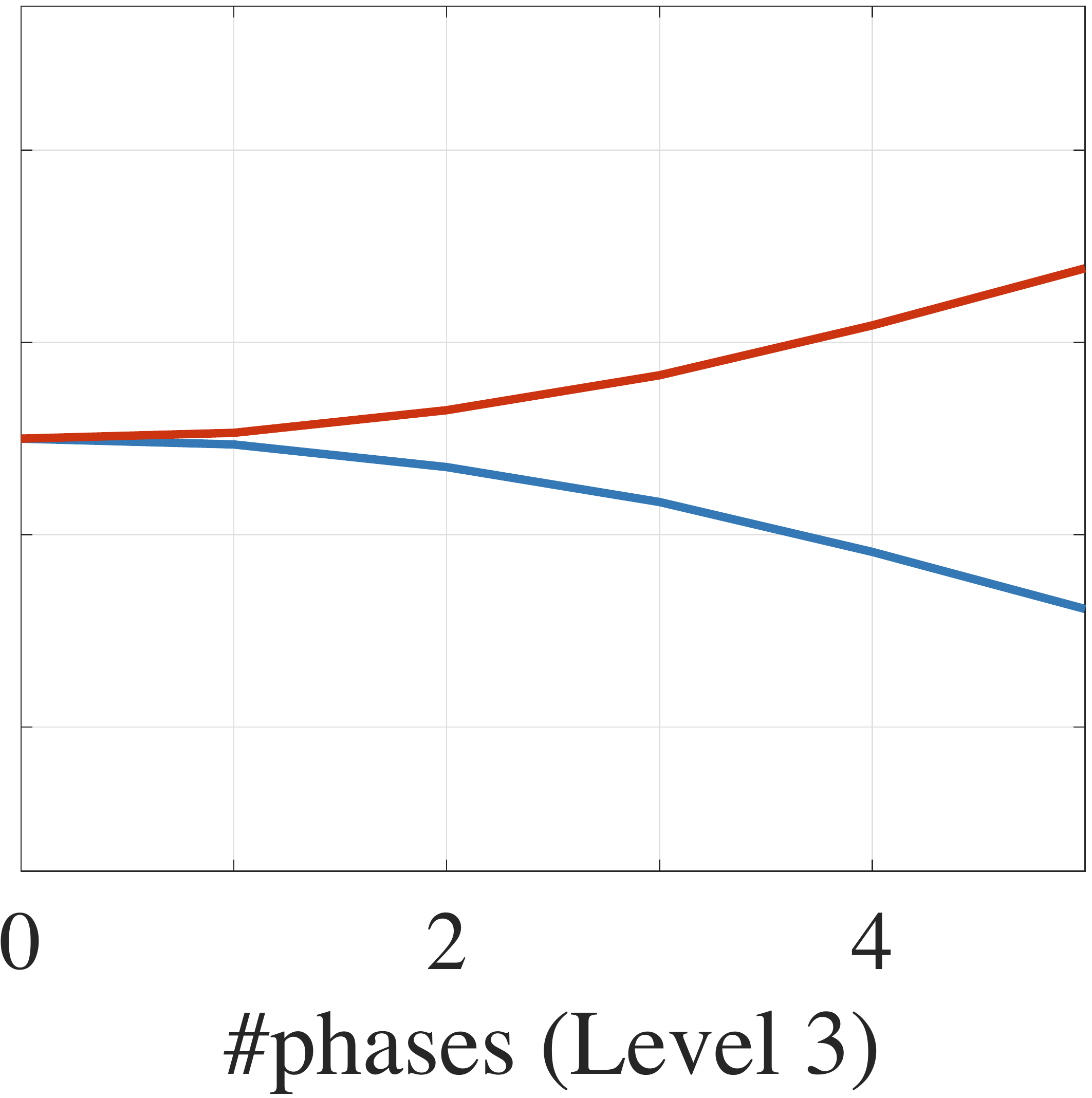}
}

\subfigure[ImageNet-Subset, $N$=25]{
\hspace{0.1cm}
\cifarincludegraphics{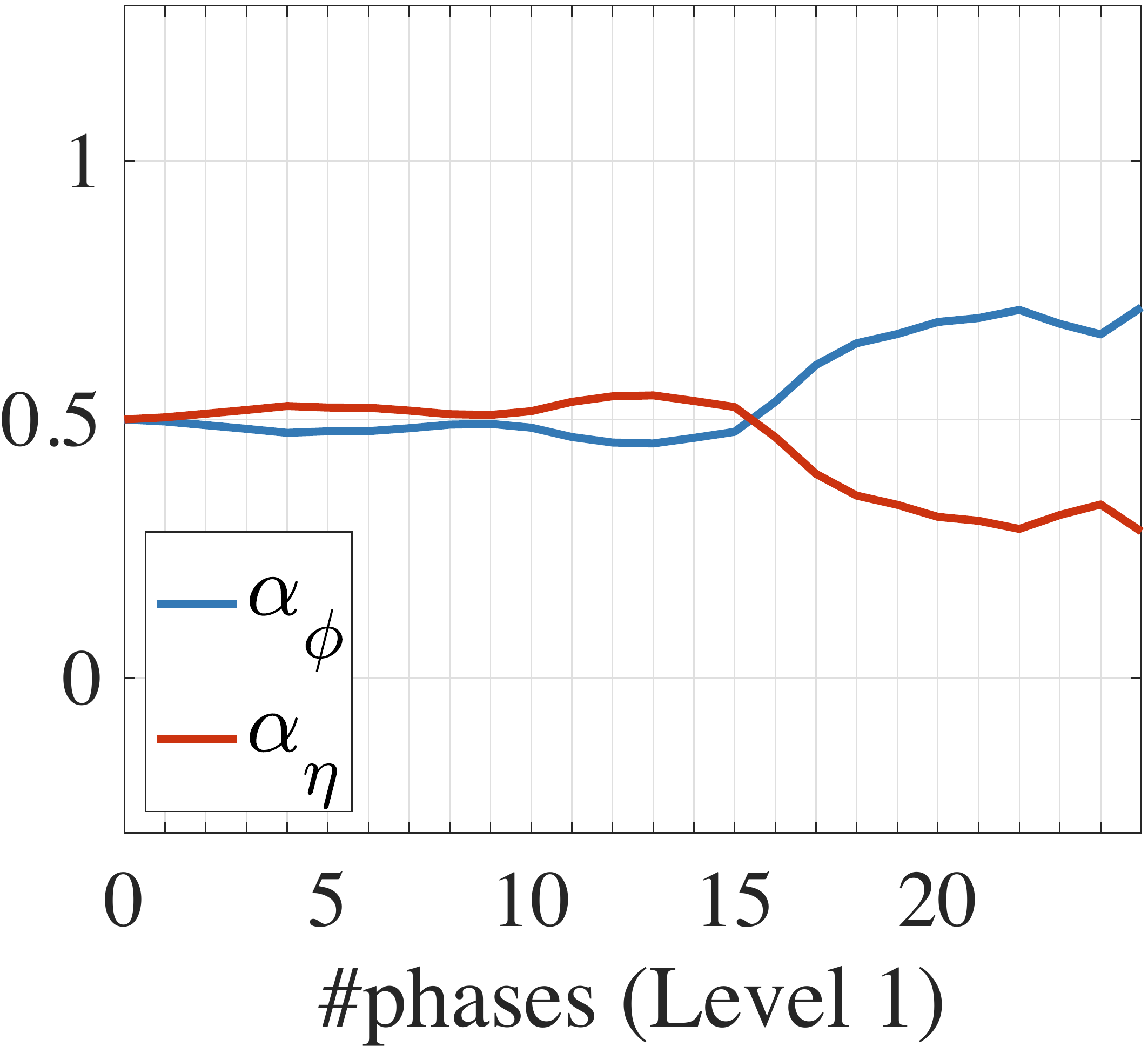}
\hspace{0.2cm}
\cifarincludegraphics{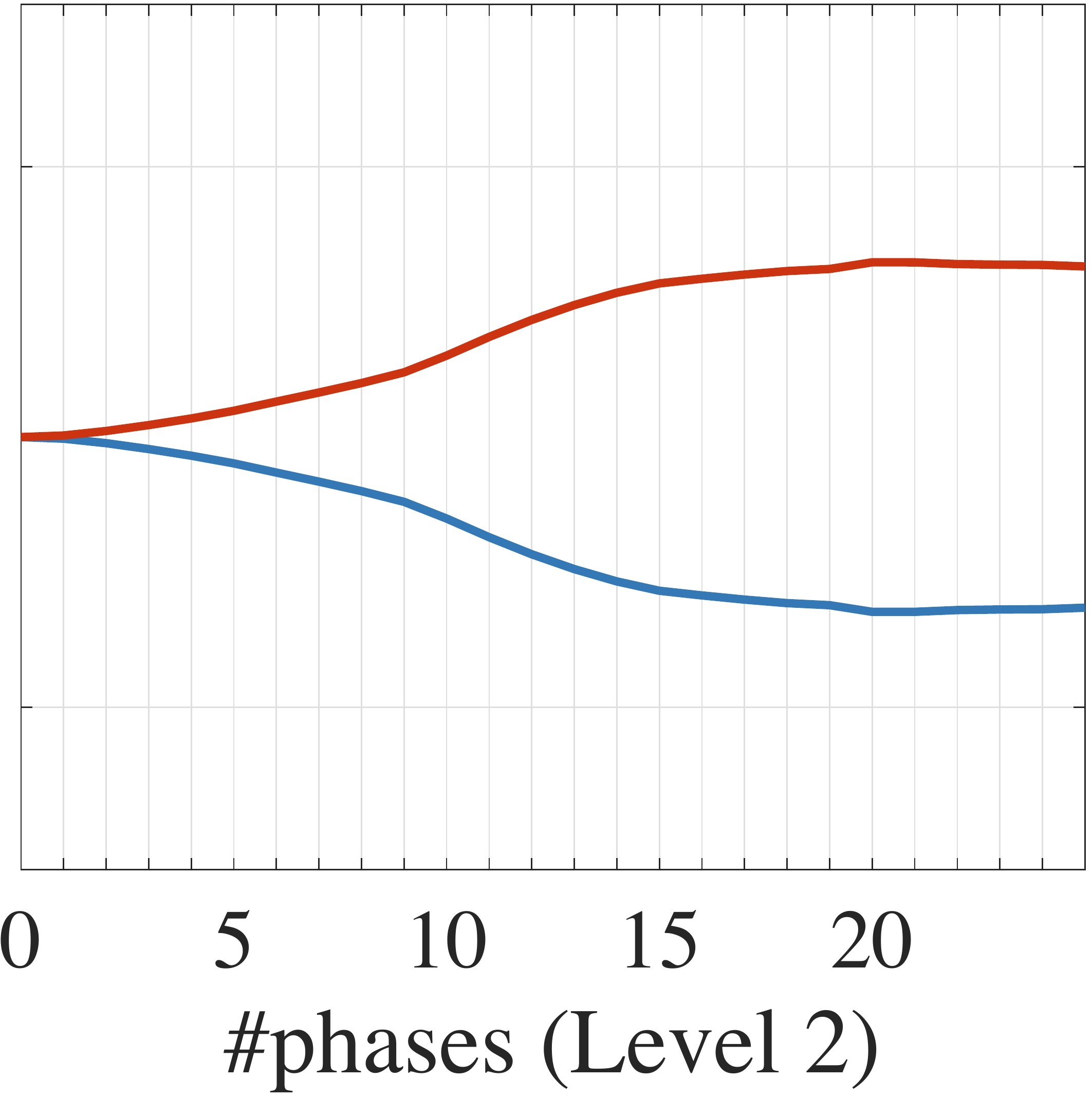}
\hspace{0.2cm}
\cifarincludegraphics{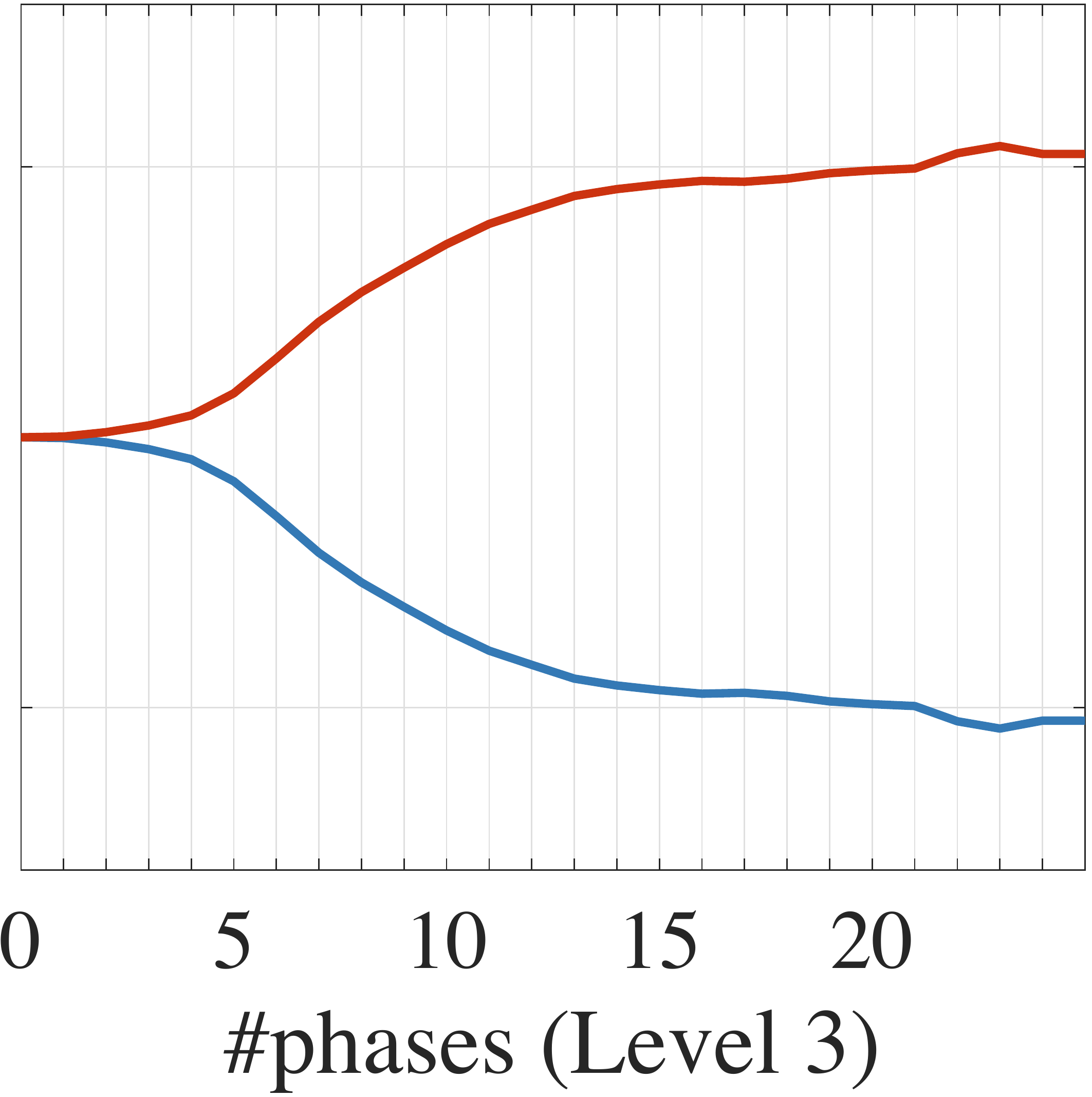}
}

\caption{\mycaptionsupp{Supplementary to Figure~\textcolor{red}{4}.} The changes of values for $\alpha_{\eta}$ and $\alpha_{\phi}$ on ImageNet-Subset.}
\label{figure_supp_values_imgnetsub}
\end{figure*}